\documentclass[accepted]{uai2025} 
                        
\usepackage[american]{babel}

\usepackage{natbib} 
\usepackage{mathtools} 
\usepackage{booktabs} 
\usepackage{tikz} 

\usepackage{tikz-cd}
\usetikzlibrary{arrows}
\usepackage{comment}
\usepackage{amssymb}
\usepackage{tikz}
\usepackage{tikz-qtree,tikz-qtree-compat}
\usetikzlibrary{positioning}
\usepackage{float}

\usepackage{amsmath}
\usepackage{amsthm}
\usepackage{tikz}
\usepackage{tikz-qtree,tikz-qtree-compat}

\usepackage{stackengine}
\def\delequal{\mathrel{\ensurestackMath{\stackon[1pt]{=}{\scriptstyle\Delta}}}}
\theoremstyle{plain}
\newtheorem{theorem}{Theorem}

\theoremstyle{definition}
\newtheorem{definition}[theorem]{Definition}

\theoremstyle{remark}

\usepackage{subcaption}

\title{Estimating Probabilities of Causation with Machine Learning Models}

\author[1]{\href{mailto:<sw23v@fsu.edu>?Subject=Your UAI 2025 paper}{Shuai Wang}}
\author[1]{\href{mailto:<angli@cs.fsu.edu>?Subject=Your UAI 2025 paper}Ang Li}
\affil[1]{%
    Dept. of Computer Science\par
    Florida State University\par
    Tallahassee, FL, USA
}

\begin{document}
\maketitle
\thispagestyle{empty} 
\begin{abstract}
Probabilities of causation play a crucial role in modern decision-making. This paper addresses the challenge of predicting probabilities of causation for subpopulations with insufficient data using machine learning models. Tian and Pearl first defined and derived tight bounds for three fundamental probabilities of causation: the probability of necessity and sufficiency (PNS), the probability of sufficiency (PS), and the probability of necessity (PN). However, estimating these probabilities requires both experimental and observational distributions specific to each subpopulation, which are often unavailable or impractical to obtain with limited population-level data. We assume that the probabilities of causation for each subpopulation are determined by its characteristics. To estimate these probabilities for subpopulations with insufficient data, we propose using machine learning models that draw insights from subpopulations with sufficient data. Our evaluation of multiple machine learning models indicates that, given sufficient population-level data and an appropriate choice of machine learning model and activation function, PNS can be effectively predicted. Through simulation studies, we show that our multilayer perceptron (MLP) model with the Mish activation function achieves a mean absolute error (MAE) of approximately 0.02 in predicting PNS for 32,768 subpopulations using data from around 2,000 subpopulations.
\end{abstract}

\section{Introduction}\label{sec:intro}
Understanding causal relationships and estimating probabilities of causation are crucial in fields such as healthcare, policy evaluation, and economics \citep{pearl2009causality, imbens2015causal, heckman2015causal}. Unlike correlation-based methods, causal inference enables decision-makers to determine whether an action or intervention directly leads to a desired outcome. This is particularly essential in personalized medicine, where accurately assessing treatment effects ensures both efficacy and safety \citep{mueller:pea23-r530}. Moreover, causal reasoning enhances machine learning applications by improving accuracy \citep{li2020training}, interpretability, and fairness \citep{plecko2022causal} in automated decision-making. Despite its broad significance, estimating probabilities of causation remains challenging due to data limitations. In this paper, we address this challenge by leveraging machine learning techniques to predict probabilities of causation for subpopulations with insufficient data.

The study of probabilities of causation began around 2000 when \cite{pearl1999probabilities} first defined three fundamental probabilities—PNS, PS, and PN—within Structural Causal Models \citep{galles1998axiomatic,halpern2000axiomatizing,pearl2009causality}. Subsequently, \cite{tian2000probabilities} derived tight bounds for these probabilities using Balke's linear programming \citep{balke1995probabilistic}, incorporating both observational and experimental data. Nearly two decades later, \cite{li2019unit} formally proved these bounds and introduced the unit selection model, a decision-making framework based on their linear combination. More recently, \cite{li2024unit} extended the definitions and bounds to a more general form. Additionally, \cite{pearl:etal21-r505}, as well as \cite{dawid2017}, demonstrated that these bounds could be further refined given specific causal structures.

However, any above estimation of the probabilities of causation requires both observational and experimental data. Additionally, estimating (sub)populations, based on \cite{li2022probabilities}'s suggestions, requires approximately $1,300$ entries of both data types for each (sub)population, making the process impractical. \cite{li2022learning,li2022unitlearning} demonstrated the potential of machine learning models to achieve accurate estimations for (sub)populations. In this research, we select five diverse machine learning models based on the characteristics of the probabilities of causation. We then evaluate their performance in accomplishing the task.

\subsection{Contributions}
Despite the extensive theoretical research on probabilities of causation, practical estimation methods have remained unexplored. Our work provides the first systematic approach to predicting probabilities of causation using machine learning. Specifically, we make the following contributions:
\begin{itemize}[nosep]
    \item \textbf{First Machine Learning Pipeline for Predicting Probabilities of Causation:} We propose a novel machine learning framework to estimate the bounds of PNS, PS, and PN, filling a critical gap between theoretical causal inference and practical applications.
    \item \textbf{First Accurate Machine Learning Model for PNS Prediction:} We demonstrate that a MLP can accurately predict PNS, proving that machine learning is a feasible and effective tool for estimating probabilities of causation.
    \item \textbf{First Dataset for PNS Bound Prediction:} We construct and release the first synthetic dataset specifically designed to evaluate machine learning models for estimating PNS, providing a foundation for future research in this area.
\end{itemize}
To the best of our knowledge, no prior work has applied machine learning to the problem of predicting probabilities of causation. Our study establishes a new research direction by bridging causal inference and machine learning for practical estimation tasks.

The remainder of the paper is structured as follows: first, we review key causal inference concepts to provide necessary context. Next, we introduce the model and dataset used in our study. Finally, we present our five machine learning models developed for the task. All code for data generation and machine learning models is included in the appendix.

\section{Preliminaries}
\label{related work}
In this section, we review the fundamental concepts of causal inference necessary for understanding the rest of the paper. We begin by discussing the definitions of PNS, PS, and PN as introduced by \cite{pearl1999probabilities}, followed by the definitions of identifiability and the conditions required to identify PNS, PS, and PN \citep{tian2000probabilities}. Additionally, we examine the tight bounds of PNS, PS, and PN in cases where they are unidentifiable \citep{tian2000probabilities}. Readers already familiar with these concepts may skip this section.

Similar to the works mentioned above, we adopt the causal language of Structural Causal Models (SCMs) \citep{galles1998axiomatic,halpern2000axiomatizing}. In this framework, the counterfactual statement ``Variable \( Y \) would have the value \( y \) had \( X \) been \( x \)'' is denoted as \( Y_x = y \), abbreviated as \( y_x \). We consider two types of data: experimental data, expressed as causal effects \( P(y_x) \), and observational data, represented by the joint probability function \( P(x, y) \). Unless otherwise specified, we assume \( X \) and \( Y \) are binary variables in a causal model \( M \), with \( x \) and \( y \) denoting the propositions \( X = \text{true} \) and \( Y = \text{true} \), respectively, and \( x' \) and \( y' \) representing their complements. For simplicity, we focus on binary variables; extensions to multi-valued cases are discussed by \cite{pearl2009causality} (p. 286, footnote 5) and \cite{li2024probabilities}.

First, the definitions of three basic probabilities of causation defined using SCM are as follow \citep{pearl1999probabilities}:

\begin{definition}[Probability of necessity (PN)]
Let $X$ and $Y$ be two binary variables in a causal model $M$, let $x$ and $y$ stand for the propositions $X=true$ and $Y=true$, respectively, and $x'$ and $y'$ for their complements. The probability of necessity is defined as the expression 
\begin{eqnarray}
\text{PN} &\delequal& P(Y_{X=false}=false|X=true,Y=true)\nonumber\\
 &\delequal&  P(y'_{x'}|x,y) \nonumber
\end{eqnarray}
\end{definition}

\begin{definition}[Probability of sufficiency (PS)]
Let $X$ and $Y$ be two binary variables in a causal model $M$, let $x$ and $y$ stand for the propositions $X=true$ and $Y=true$, respectively, and $x'$ and $y'$ for their complements. The probability of sufficiency is defined as the expression
\begin{eqnarray}
\text{PS} &\delequal& P(Y_{X=true}=true|X=false,Y=false)\nonumber\\
&\delequal& P(y_x|x',y') \nonumber
\end{eqnarray}
\end{definition}

\begin{definition}[Probability of necessity and sufficiency (PNS)] Let $X$ and $Y$ be two binary variables in a causal model $M$, let $x$ and $y$ stand for the propositions $X=true$ and $Y=true$, respectively, and $x'$ and $y'$ for their complements. The probability of necessity and sufficiency is defined as the expression
\begin{eqnarray}
\text{PNS} &\delequal& P(Y_{X=true}=true,Y_{X=false}=false)\nonumber\\
&\delequal& P(y_x,y'_{x'}) \nonumber
\end{eqnarray}
\end{definition}
Then, we review the identification conditions for PNS, PS, and PN \citep{tian2000probabilities}.

\begin{definition} (Monotonicity)
A Variable $Y$ is said to be monotonic relative to variable $X$ in a causal model $M$ iff
\begin{eqnarray*}
y'_x\land y_{x'}=\text{false}.
\end{eqnarray*}
\end{definition}

\begin{theorem}
If $Y$ is monotonic relative to $X$, then PNS, PN, and PS are all identifiable, and
\begin{eqnarray*}
PNS = P(y_x) - P(y_{x'}),\\
PN = \frac{P(y) - P(y_{x'})}{P(x,y)},\\
PS = \frac{P(y_x) - P(y)}{P(x', y')}.
\end{eqnarray*}
\end{theorem}

If PNS, PN, and PS are not identifiable, informative bounds are given by \cite{tian2000probabilities}.




\begin{eqnarray}
\max \left \{
\begin{array}{cc}
0, \\
P(y_x) - P(y_{x'}), \\
P(y) - P(y_{x'}), \\
P(y_x) - P(y)
\end{array}
\right \} \le
\text{PNS}\label{pnslb}\\
\min \left \{
\begin{array}{cc}
 P(y_x), \\
 P(y'_{x'}), \\
P(x,y) + P(x',y'), \\
P(y_x) - P(y_{x'}) +\\
P(x, y') + P(x', y)
\end{array} 
\right \}\ge
\text{PNS}
\label{pnsub}\\
\max \left \{
\begin{array}{cc}
0, \\
\frac{P(y)-P(y_{x'})}{P(x,y)}
\end{array} 
\right \} \le
\text{PN} \label{pnlb}\\
\min \left \{
\begin{array}{cc}
1, \\
\frac{P(y'_{x'})-P(x',y')}{P(x,y)}
\end{array}
\right \}\ge \text{PN}
\label{pnub}\\
\max \left \{
\begin{array}{cc}
0, \\
\frac{P(y')-P(y'_{x})}{P(x',y')}
\end{array} 
\right \} \le
\text{PS} \label{pslb}\\
\min \left \{
\begin{array}{cc}
1, \\
\frac{P(y_{x})-P(x,y)}{P(x',y')}
\end{array}
\right \} \ge \text{PS}
\label{psub}
\end{eqnarray}

Therefore, the primary objective of this paper is then to predict Equations \ref{pnslb} to \ref{psub} (i.e., the lower and upper bounds of the PNS, PS, and PN) for any (sub)populations using those with sufficient data (i.e., sufficient data to estimate the distributions $P(X,Y)$ and $P(Y_X)$.) Due to space constraints, the focus will be on the bounds of PNS (i.e., Equations \ref{pnslb} and \ref{pnsub}). Unless otherwise specified, the discussion will be limited to binary treatment and effect, meaning both $X$ and $Y$ are binary.

\section{Structural Causal Model}
\label{scm}
In general, the equations in SCMs are in implicitly form (e.g., $Z=f_Z(X,Y,U_Z)$). However, in order to verify the accuracy of the learned bounds of PNS, we need to explicitly define the SCM and the data-generating process to determine the true PNS value and its bounds. Followed the setup in \cite{li2022learning}, we will use the following SCM. 
\begin{eqnarray*}
    \begin{cases}
        Z_i &= U_{Z_i} \text{ for } i \in \{1,...,20\},\\
        X&=f_X(M_X,U_X)\\
        &=\begin{cases}
            1& \text{ if } M_X+U_X > 0.5\\
            0& \text{ otherwise, }\\
        \end{cases}\\
        Y&=f_Y(X,M_Y,U_Y)\\
        &=\begin{cases}
            1& \text{ if } 0<C_Y \cdot X+M_Y+U_Y <1 \\
            1& \text{ if } 1<C_Y \cdot X+M_Y+U_Y <2 \\
            0& \text{ otherwise. }\\
        \end{cases}
    \end{cases}
\end{eqnarray*}
where $X,Y,Z_i$ are all binary, $U_{Z_i}, U_X, U_Y$ are binary exogenous variables with Bernoulli distributions, $C_Y$ is a constant, and $M_X, M_Y$ are linear combinations of $Z_i$. The randomly generated value of $C_Y,M_X,M_Y$ and the distributions of $U_X,U_Y,U_{Z_i}$ for the model are provided in the appendix.

\section{Data Generating Process}
Based on the defined model, $20$ binary features are considered (i.e., $Z_1,...,Z_{20}$). We made $15$ observable ($Z_1,...,Z_{15}$) and $5$ unobservable, and the exogenous variables are also unobservable, leading to $2^{15}$ observed subpopulations (i.e., the combination of $Z_1,...,Z_{15}$ defined a subpopulation).

\subsection{Informer Data}  
To evaluate the learned bounds, the informer data must have access to the actual PNS bounds for each subpopulation. Given the explicit form of the SCM and the distributions of all exogenous variables, the PNS bounds, as well as the experimental and observational distributions, can be computed for each combination of the features \( Z_1, \dots, Z_{15} \) (i.e., a subpopulation) using the SCM. For detailed mathematical formulations, refer to the appendix.

\subsection{Sample Collection}
A total of $50,000,000$ experimental and $50,000,000$ observational samples were generated as follows for each sample: In both settings, the exogenous variables \( U_X \), \( U_Y \), and \( U_{Z_i} \) were randomly generated according to their distributions specified in Section \ref{scm}. In the experimental setting, \( X \) was then assigned according to a \( \text{Bernoulli}(0.5) \) distribution, while \( Y \) and \( Z_i \) were computed using the structural functions described in Section \ref{scm}. In the observational setting, \( X \), \( Y \), and \( Z_i \) were all determined by the structural functions. The final datasets include only the observable features \( Z_1, \dots, Z_{15} \), along with \( X \) and \( Y \), while \( Z_{16}, \dots, Z_{20} \) were masked.

\subsection{Data for Machine Learning Models}  
We selected subpopulations from the $2^{15}$ possible groups that contained at least $1,300$ experimental and observational samples ($1,300$ based on \cite{li2022probabilities}'s suggestions). For these selected subpopulations, we computed the experimental and observational distributions and determined the bounds of PNS using Equations \ref{pnslb} and \ref{pnsub}. These results served as the data for our machine learning models (i.e., each data entry consists of 15 features and the PNS bounds as the label.) The obtained data includes $2,054$ entries for the lower bound (LB) and $2,065$ entries for the upper bound (UB) of the PNS.
\section{Machine Learning Prediction}\label{sec:ml}
\begin{figure*}[!htbp] 
    \centering
    \includegraphics[width=\textwidth]{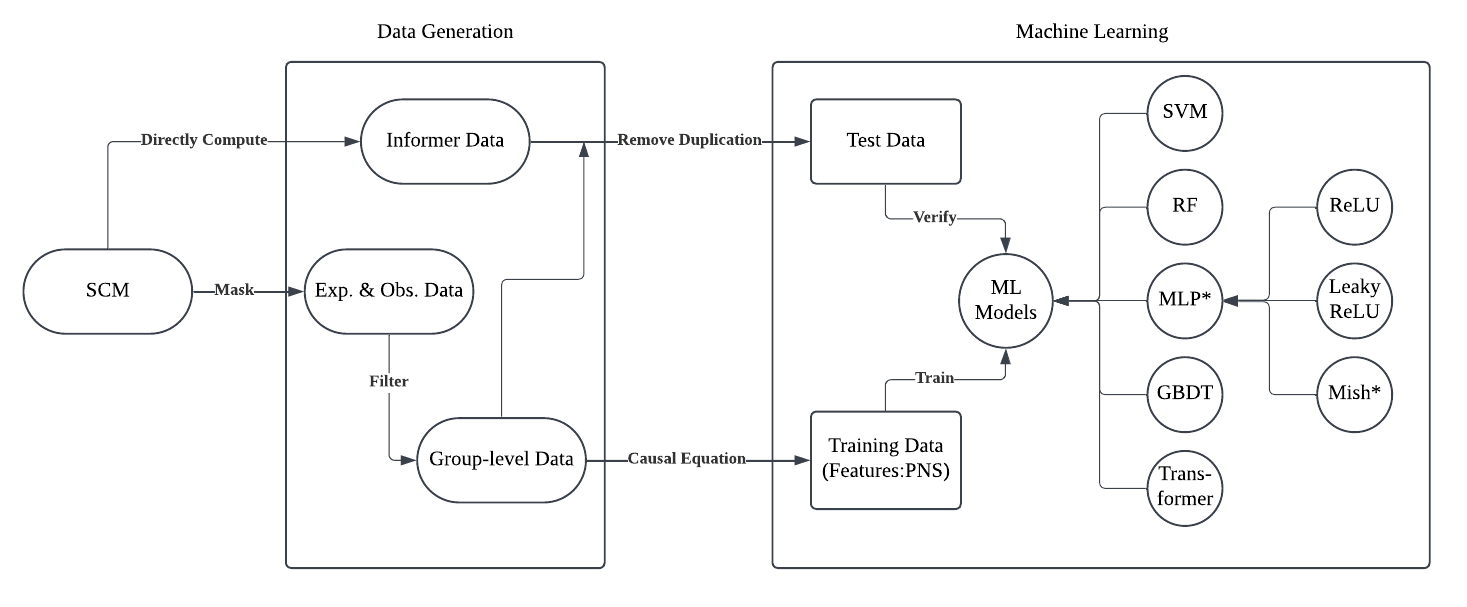}
    \caption{Framework for Causal Data Generation and Machine Learning Prediction.}
    \label{fig:flow chart}
\end{figure*}

To evaluate the feasibility of machine learning in predicting the bounds of the PNS, we employed five distinct machine learning models to assess their effectiveness in this task: Support Vector Machine (SVM) \citep{svm}, Random Forest (RF) \citep{rf}, Gradient Boosting Decision Trees (GBDT) \citep{gbdt}, Transformer \citep{vaswani2017attention}, and Multilayer Perceptron (MLP) \citep{mlp}. These models were chosen to represent a diverse range of machine learning paradigms, including kernel-based methods (SVM), ensemble learning techniques (RF and GBDT), and deep learning approaches (MLP and Transformer). This selection ensures a comprehensive evaluation of their ability to approximate causal quantities across different settings. A detailed pipeline is illustrated in Figure \ref{fig:flow chart}.

\begin{figure*}[!htb]
    \centering
    \begin{subfigure}[b]{0.24\linewidth}
        \centering
        \includegraphics[width=\linewidth]{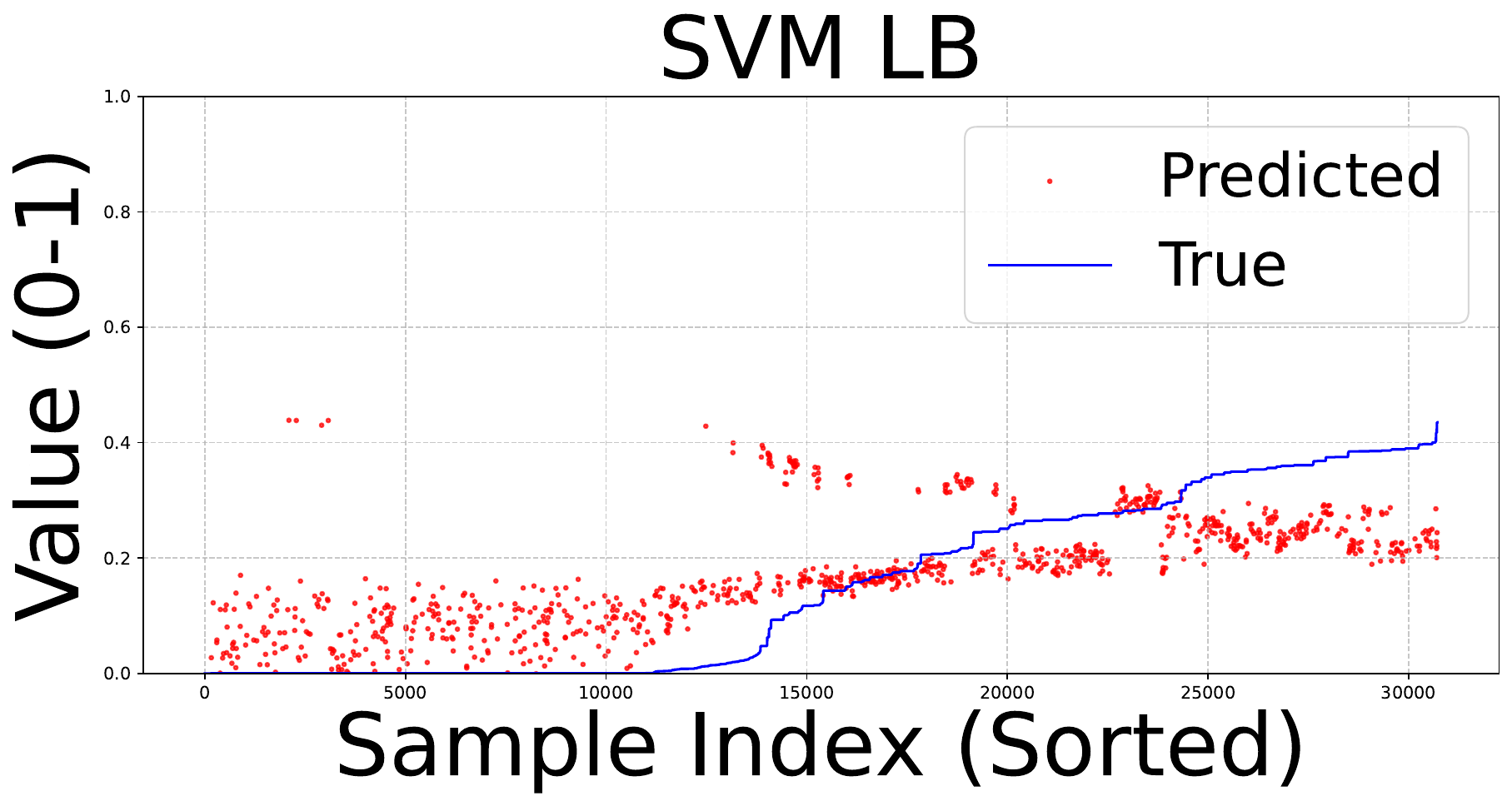}
        \caption{SVM (Lower bound)}
        \label{fig:svm2}
    \end{subfigure}
    \hfill
    \begin{subfigure}[b]{0.24\linewidth}
        \centering
        \includegraphics[width=\linewidth]{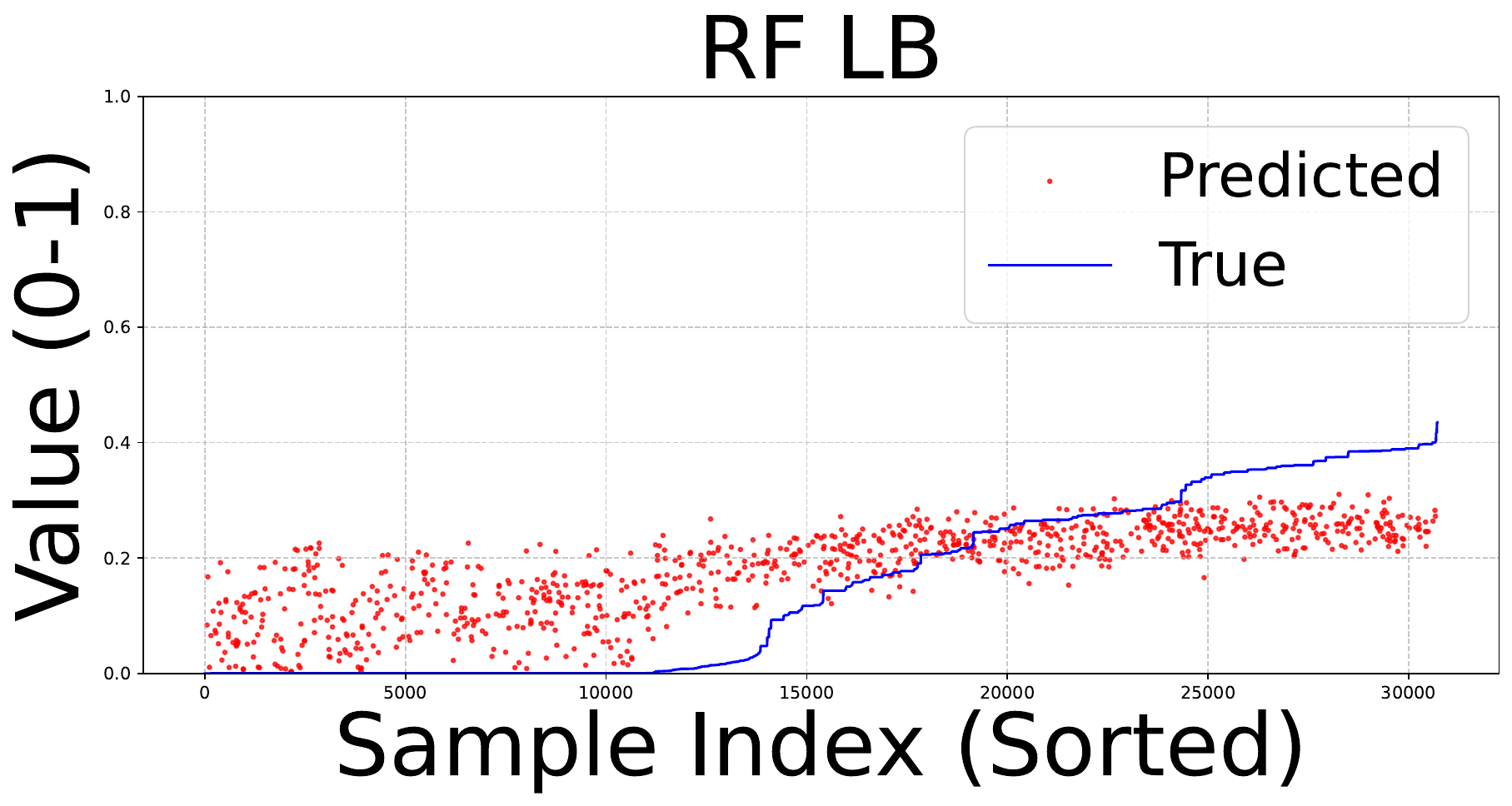}
        \caption{RF (Lower bound)}
        \label{fig:rf2}
    \end{subfigure}
    \hfill
    \begin{subfigure}[b]{0.24\linewidth}
        \centering
        \includegraphics[width=\linewidth]{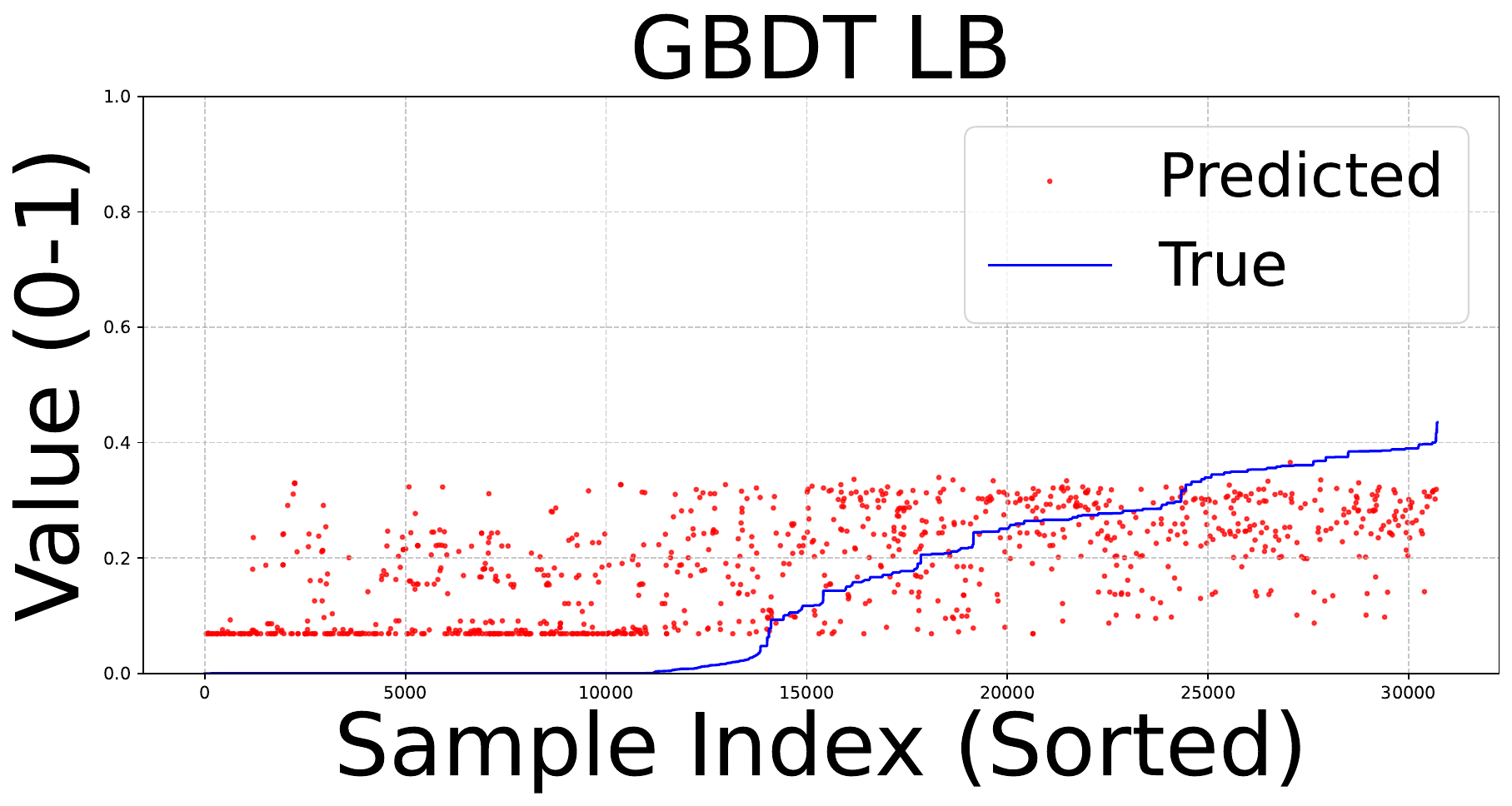}
        \caption{GBDT (Lower bound)}
        \label{fig:gbdt2}
    \end{subfigure}
    \hfill
    \begin{subfigure}[b]{0.24\linewidth}
        \centering
        \includegraphics[width=\linewidth]{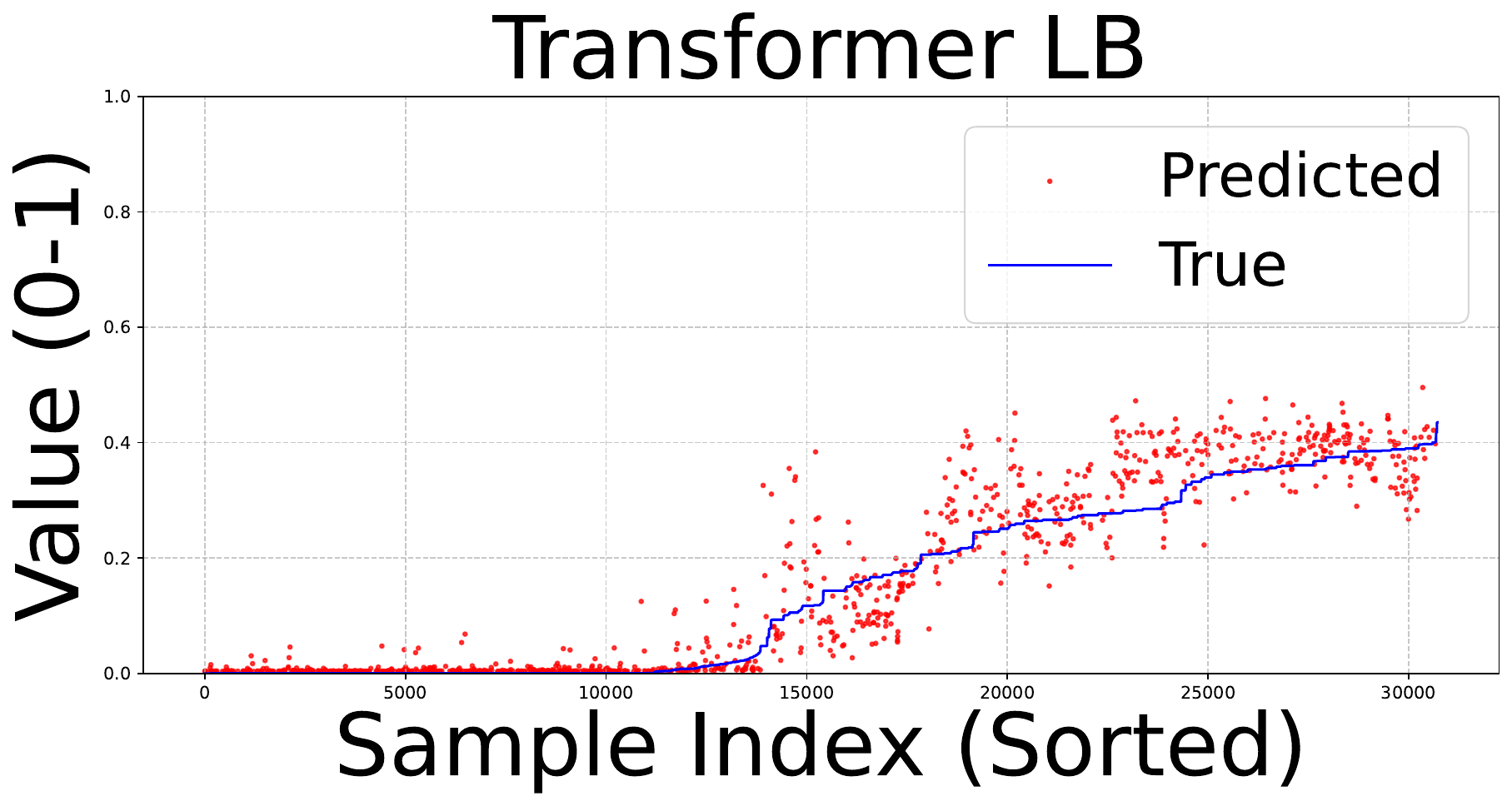}
        \caption{Transformer (Lower bound)}
        \label{fig:transformer2}
    \end{subfigure}

    \vspace{0.3cm} 

    \begin{subfigure}[b]{0.24\linewidth}
        \centering
        \includegraphics[width=\linewidth]{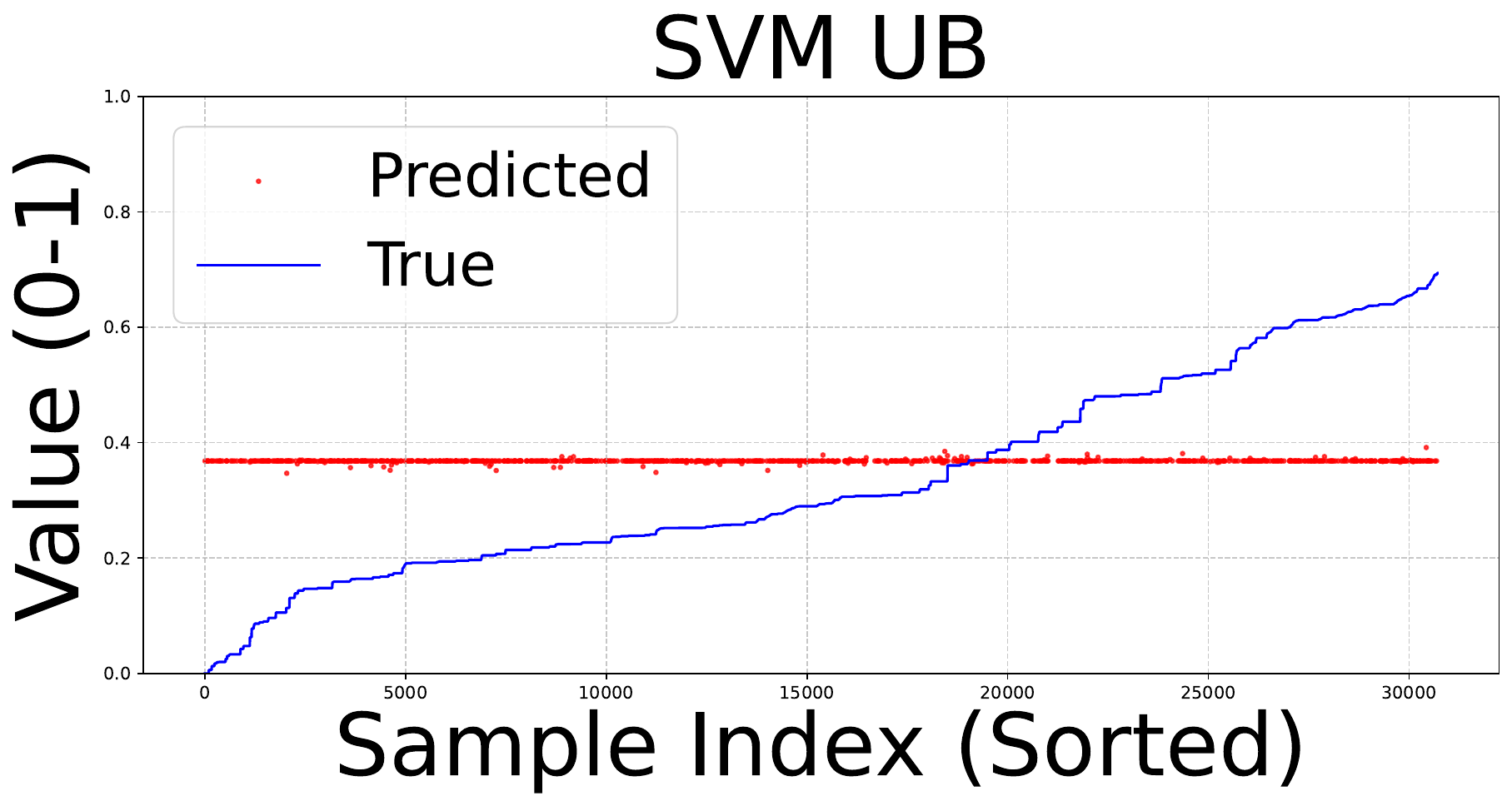}
        \caption{SVM (Upper bound)}
        \label{fig:svm3}
    \end{subfigure}
    \hfill
    \begin{subfigure}[b]{0.24\linewidth}
        \centering
        \includegraphics[width=\linewidth]{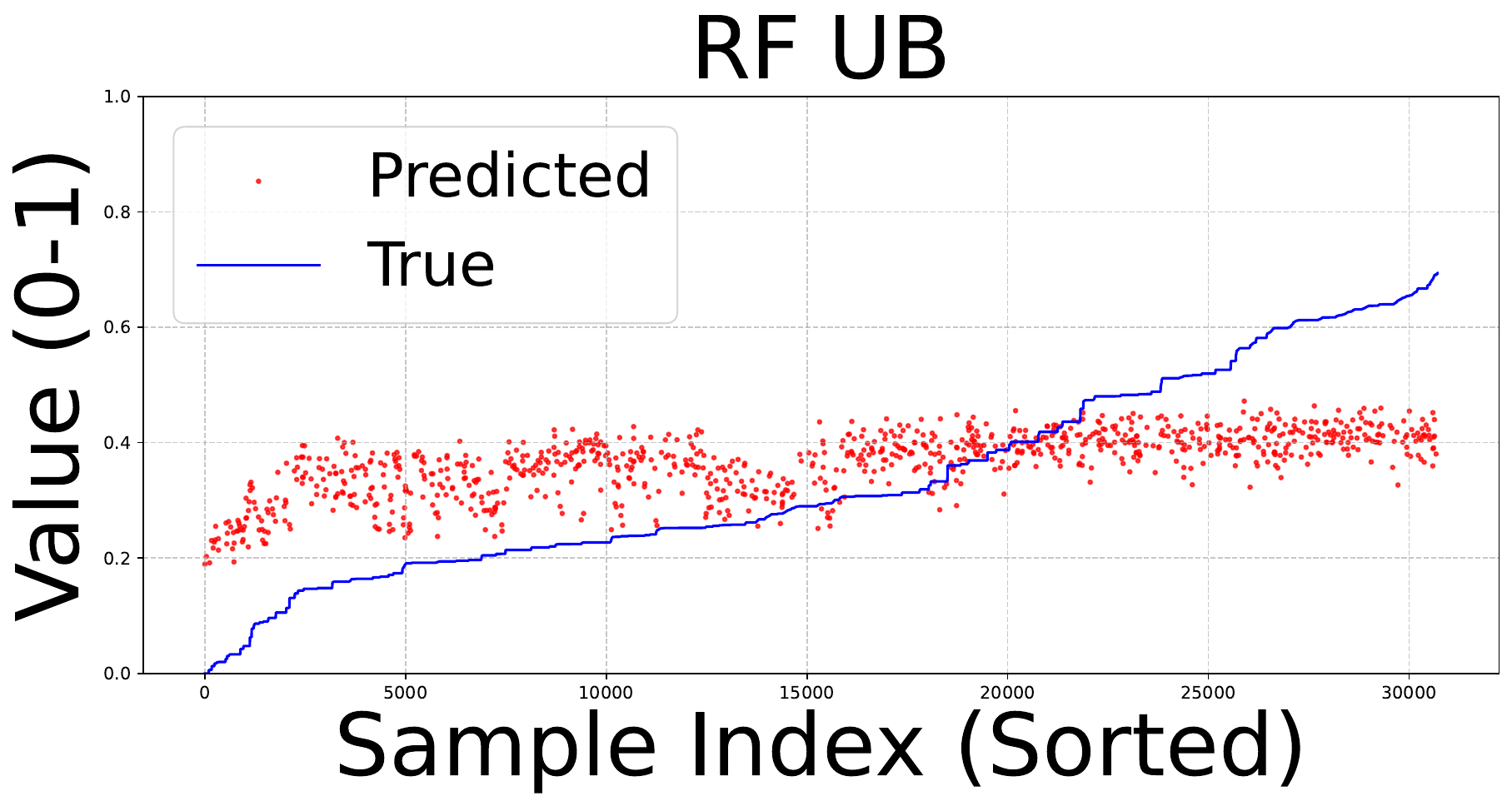}
        \caption{RF (Upper bound)}
        \label{fig:rf3}
    \end{subfigure}
    \hfill
    \begin{subfigure}[b]{0.24\linewidth}
        \centering
        \includegraphics[width=\linewidth]{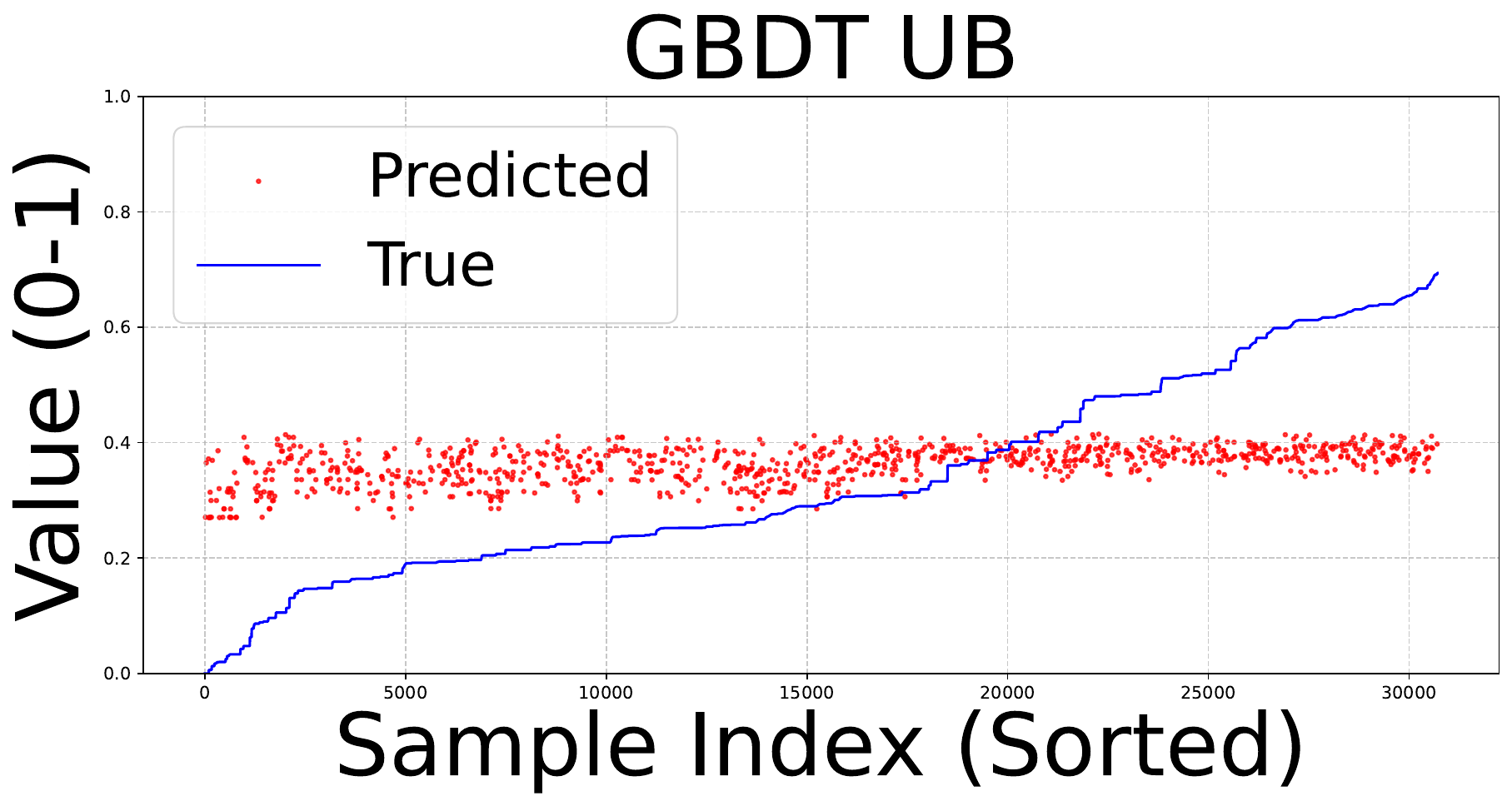}
        \caption{GBDT (Upper bound)}
        \label{fig:gbdt3}
    \end{subfigure}
    \hfill
    \begin{subfigure}[b]{0.24\linewidth}
        \centering
        \includegraphics[width=\linewidth]{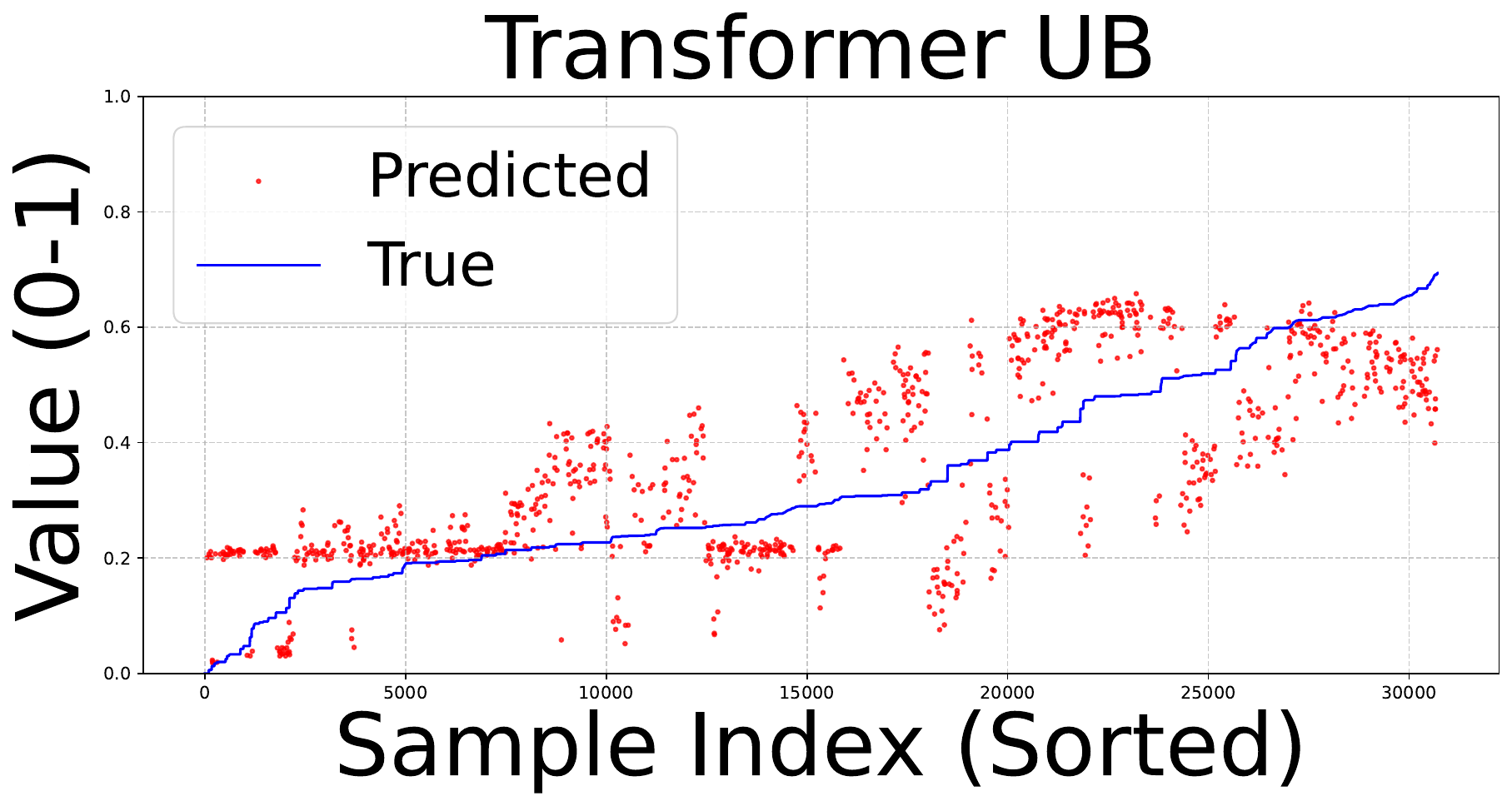}
        \caption{Transformer (Upper bound)}
        \label{fig:transformer3}
    \end{subfigure}

    \caption{Comparison of true and predicted values across different models for both lower and upper bounds.}
    \label{fig:combined_model_comparison}
\end{figure*}
\subsection{Support Vector Machine}
Support Vector Machines (SVM) \citep{svm} are widely used and well-established supervised learning models. Given their strengths, we selected Support Vector Regression (SVR), a variant of SVM, as the first model for our experiments. To effectively capture complex patterns, we employed the Radial Basis Function (RBF) kernel to map the data into a high-dimensional feature space.

Key hyperparameters include the penalty parameter (\( C \)), the insensitive loss threshold (\( \epsilon \)), and the kernel coefficient (\( \gamma \)). The parameter \( C \) controls the trade-off between model complexity and error tolerance, where larger values may lead to overfitting. The threshold \( \epsilon \) defines the margin of tolerance for errors, while \( \gamma \) determines the influence range of individual data points.

A two-stage hyperparameter tuning strategy was adopted. First, Randomized Search \citep{randomsearch} was employed to efficiently explore the parameter space and identify promising ranges. Then, Grid Search \citep{randomsearch} was used to fine-tune parameters within these ranges. Cross-validation ensured robust generalization throughout the tuning process.

\begin{figure*}[!htb]
    \centering
    \begin{subfigure}[b]{0.32\linewidth}
        \centering
        \includegraphics[width=\linewidth]{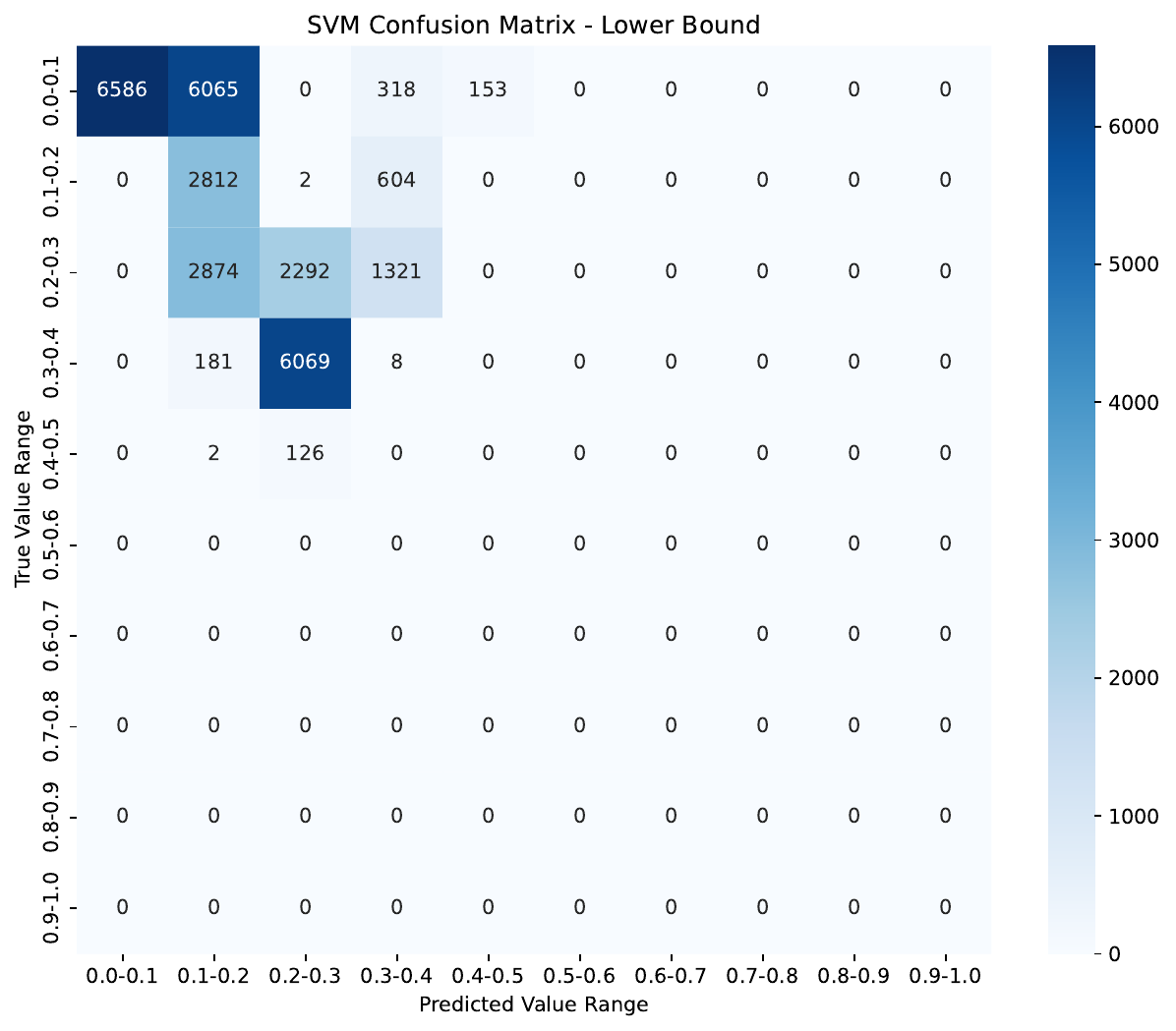}
        \caption{SVM Lower bound.}
        \label{fig:svm_lb} 
    \end{subfigure}
    \hfill
    \begin{subfigure}[b]{0.32\linewidth}
        \centering
        \includegraphics[width=\linewidth]{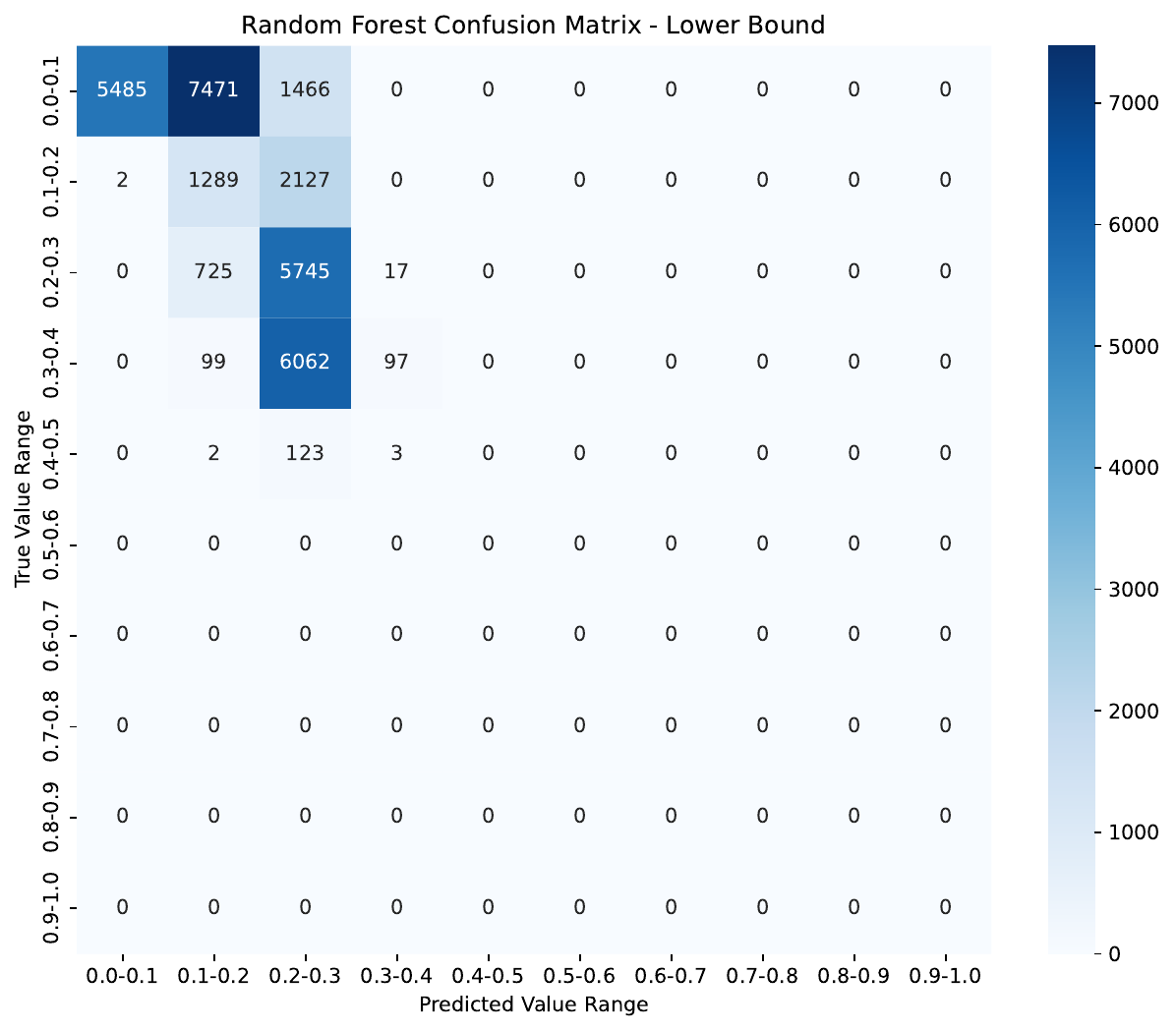}
        \caption{RF Lower bound.}
        \label{fig:rf_lb}
    \end{subfigure}
    \hfill
    \begin{subfigure}[b]{0.32\linewidth}
        \centering
        \includegraphics[width=\linewidth]{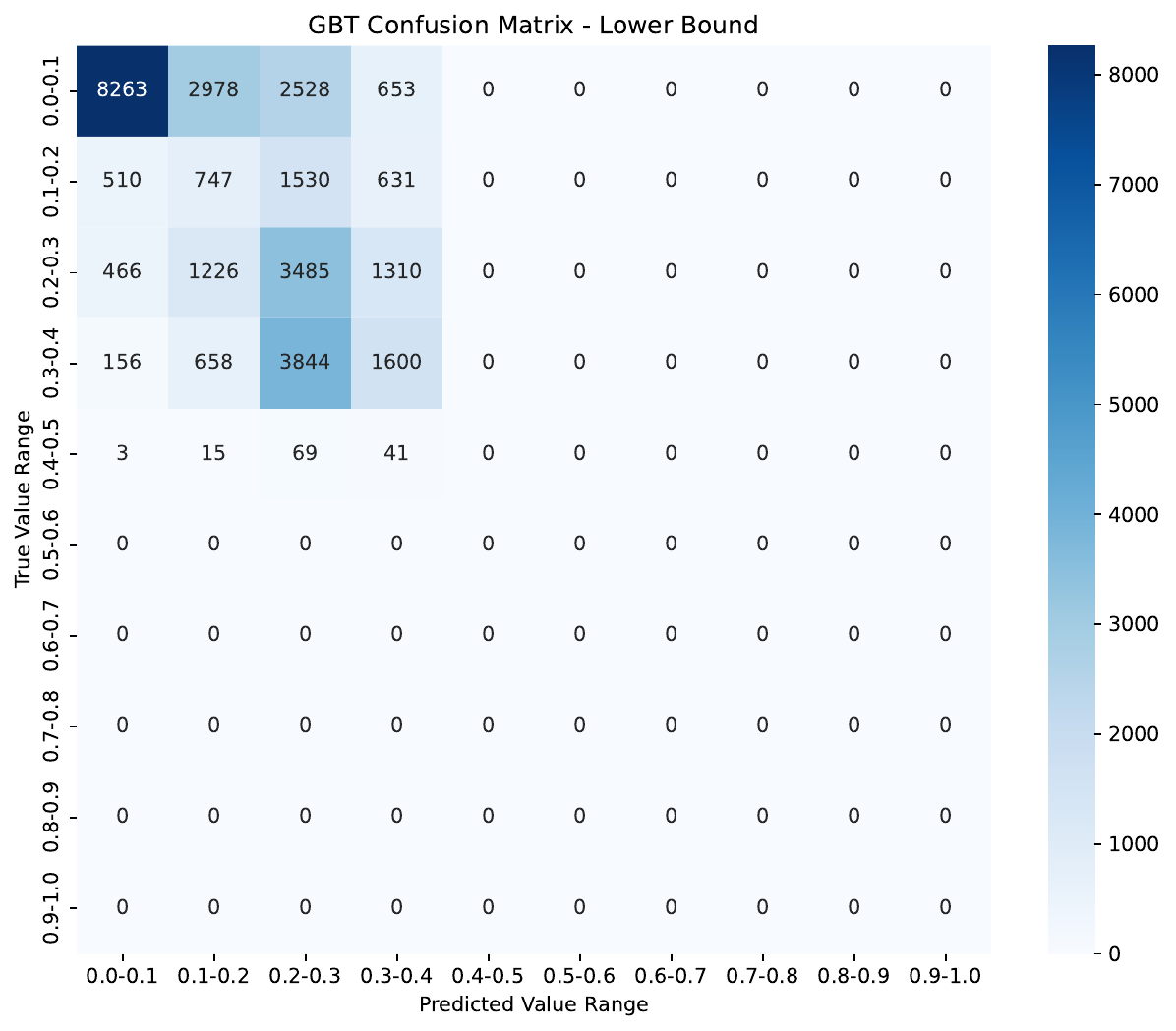}
        \caption{GBDT Lower bound.}
        \label{fig:gbdt_lb}
    \end{subfigure}

    \begin{subfigure}[b]{0.32\linewidth}
        \centering
        \includegraphics[width=\linewidth]{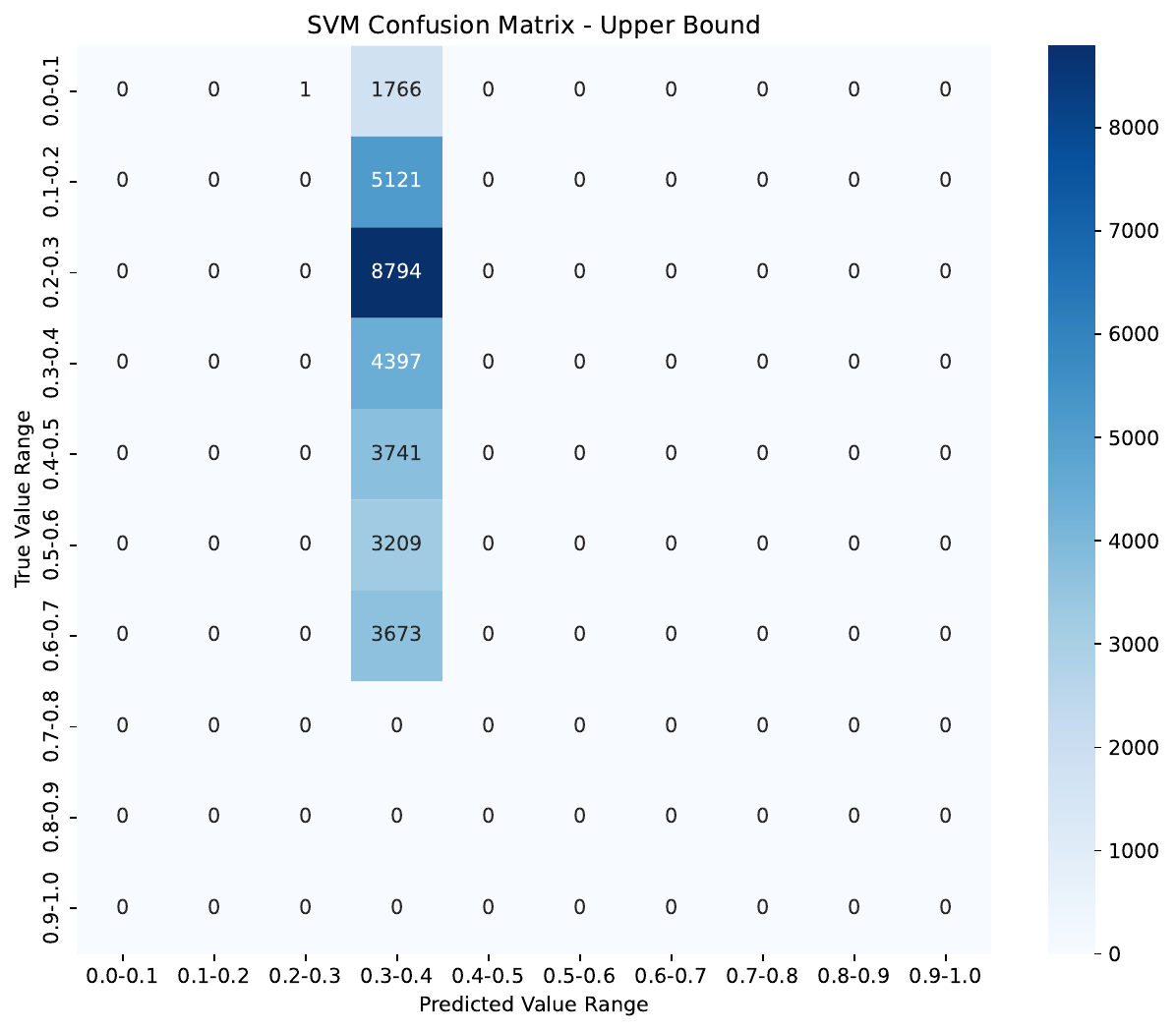}
        \caption{SVM Upper bound.}
        \label{fig:svm_ub}
    \end{subfigure}
    \hfill
    \begin{subfigure}[b]{0.32\linewidth}
        \centering
        \includegraphics[width=\linewidth]{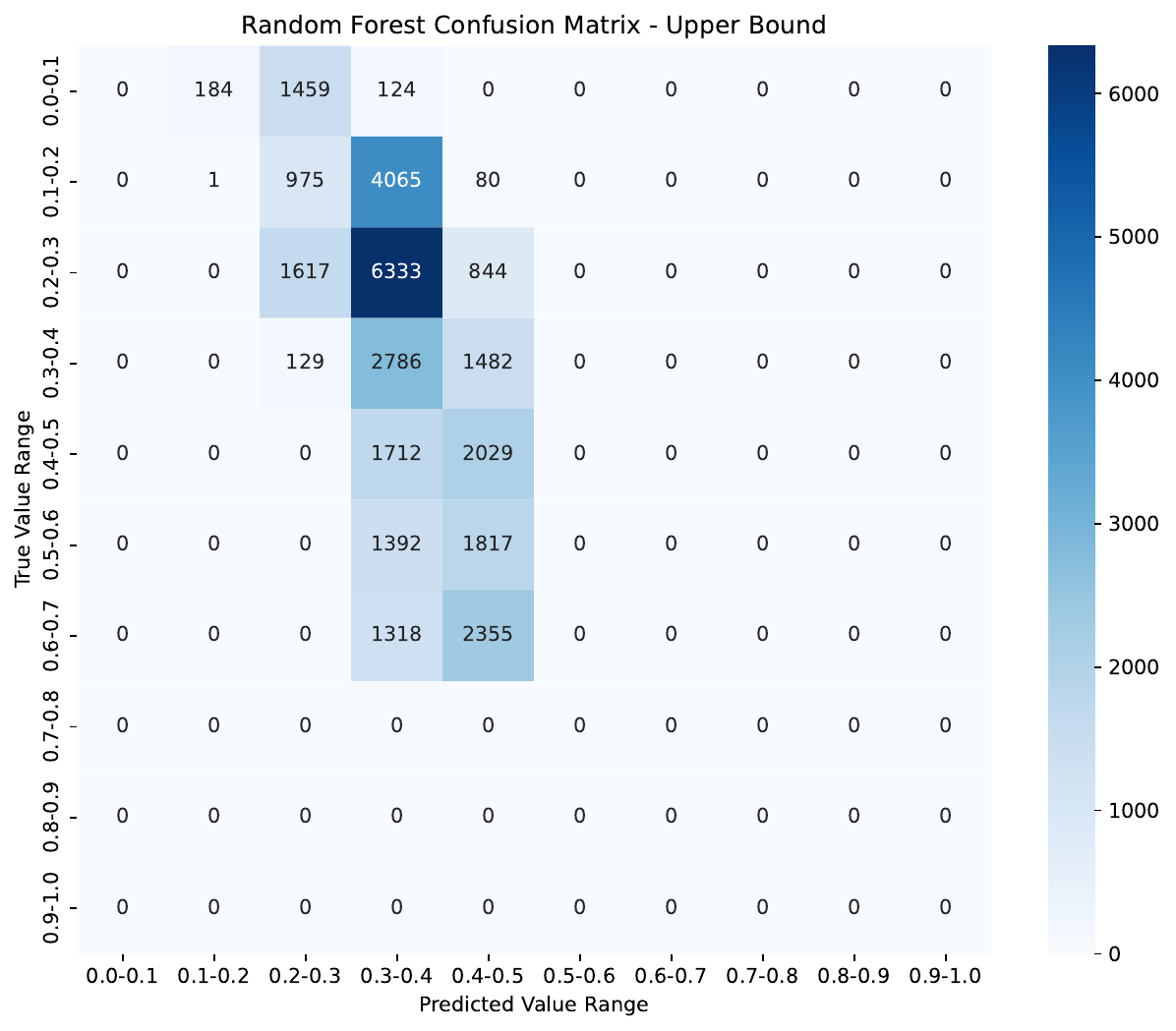}
        \caption{RF Upper bound.}
        \label{fig:rf_ub}
    \end{subfigure}
    \hfill
    \begin{subfigure}[b]{0.32\linewidth}
        \centering
        \includegraphics[width=\linewidth]{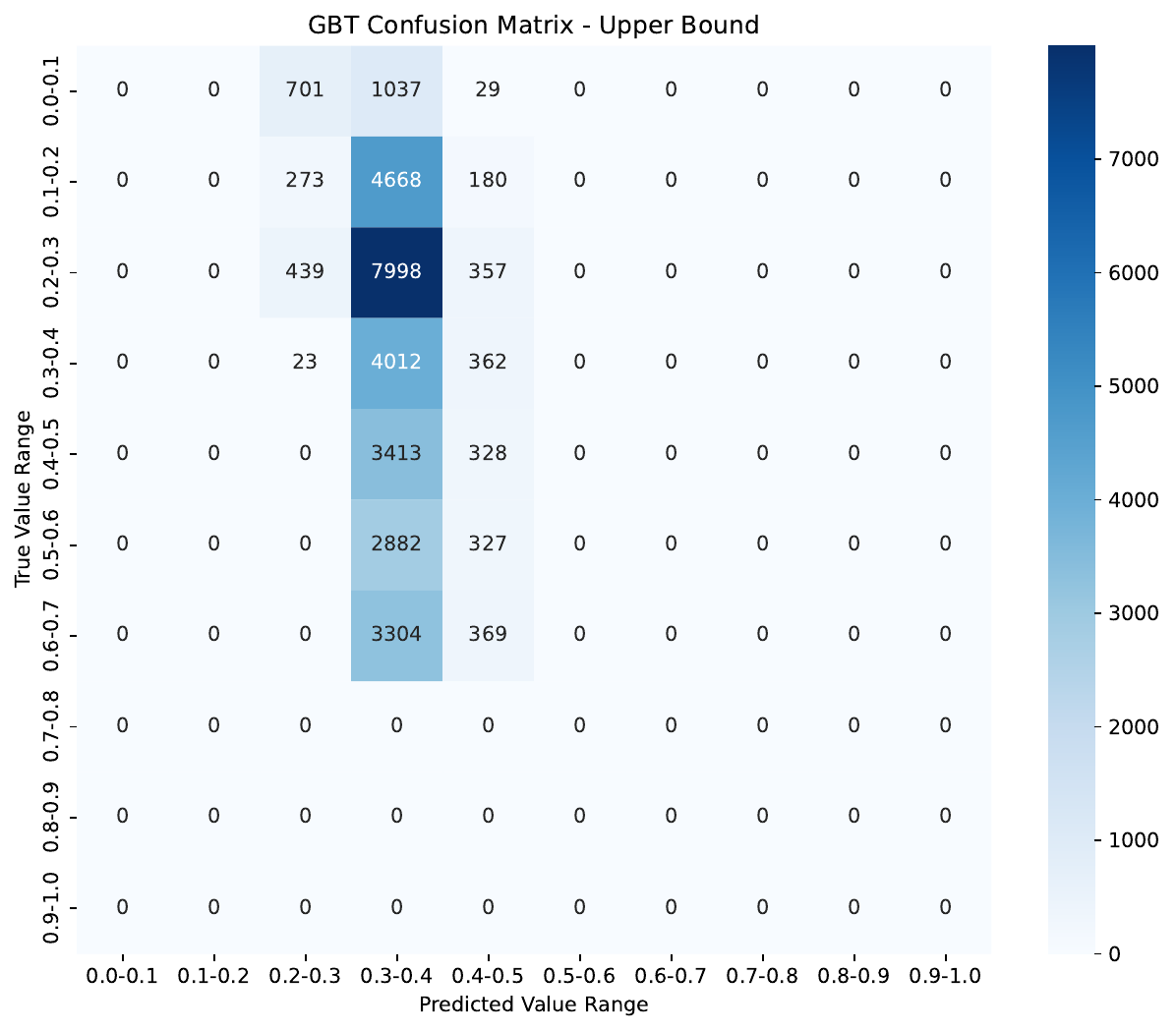}
        \caption{GBDT Upper bound.}
        \label{fig:gbdt_ub}
    \end{subfigure}

    \begin{subfigure}[b]{0.32\linewidth}
        \centering
        \includegraphics[width=\linewidth]{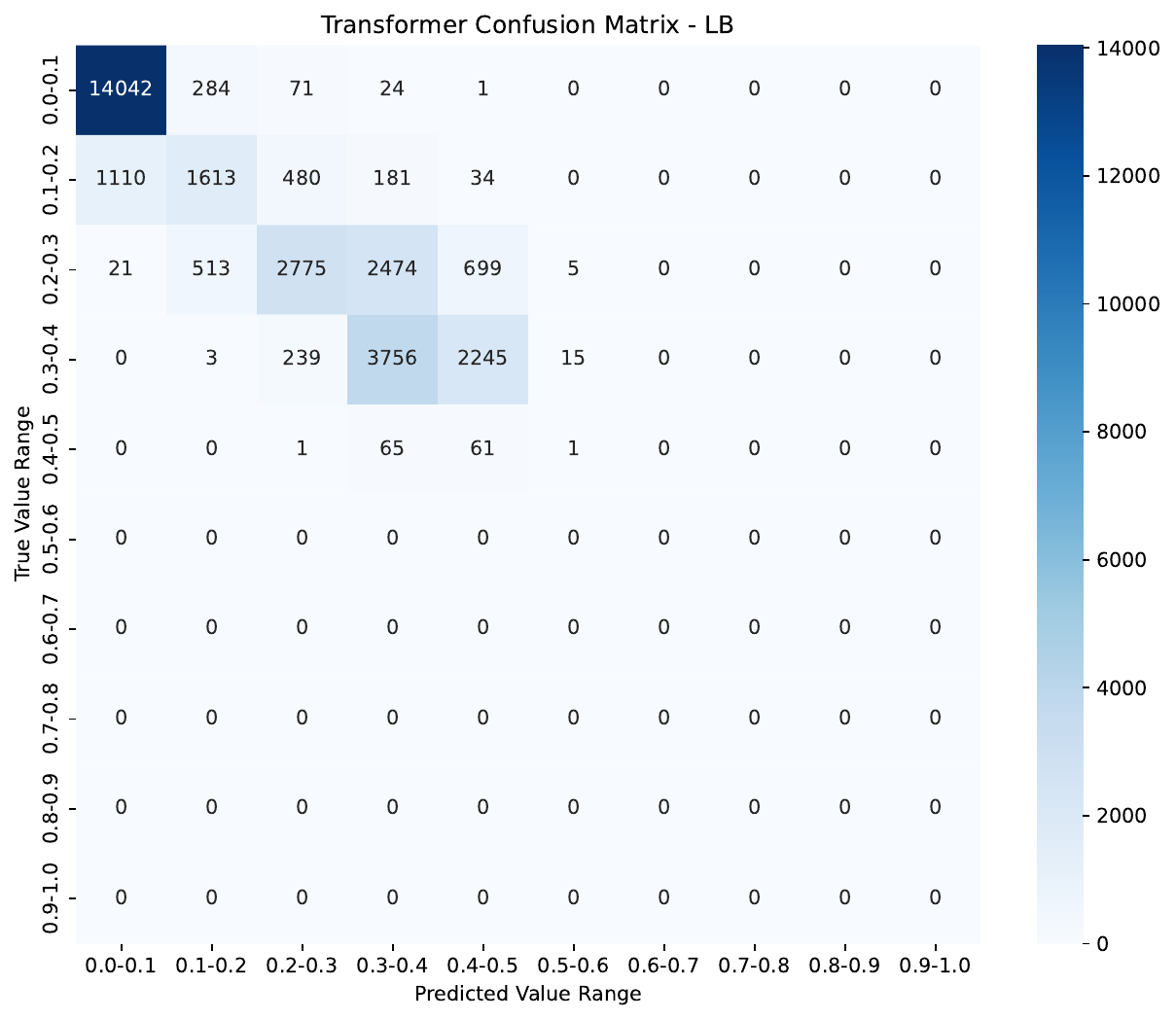}
        \caption{Transformer Lower bound.}
        \label{fig:transformer_lb}
    \end{subfigure}
    \begin{subfigure}[b]{0.32\linewidth}
        \centering
        \includegraphics[width=\linewidth]{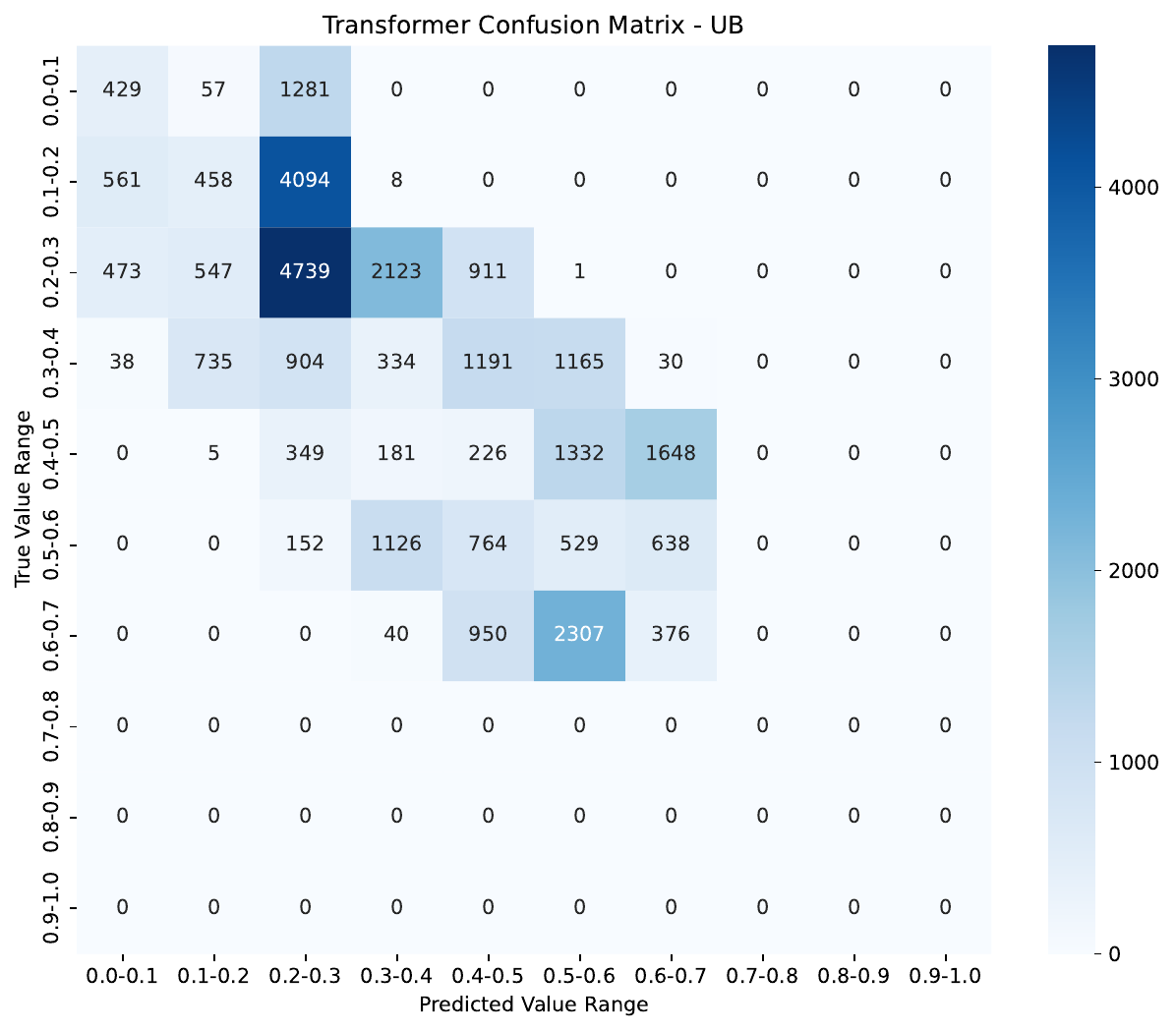}
        \caption{Transformer Upper bound.}
        \label{fig:transformer_ub}
    \end{subfigure}

    \caption{Confusion matrices of SVM, RF, GBDT, and Transformer models.}
    \label{fig:confusion_matrices_combined}
\end{figure*}

Finally, the mean squared error (MSE) and mean absolute error (MAE) values of the SVR model can be found in Table \ref{tab:comparison}. Confusion matrices are presented in Figures \ref{fig:svm_lb} and \ref{fig:svm_ub}, while Figures \ref{fig:svm2} and \ref{fig:svm3} provide a clearer comparison with the true PNS bounds. For the prediction of the lower bound, SVR demonstrates reasonable effectiveness; however, for the more complex upper bound, it exhibits a significant decline in accuracy.

\subsection{Random Forest}
Random Forests (RF) \citep{rf} are a widely used ensemble learning method for classification, regression, and other predictive tasks. The core idea behind RF is to construct multiple decision trees during training and aggregate their outputs to enhance overall performance. As an ensemble model, RF exhibits strong robustness, motivating us to assess its effectiveness in predicting PNS bounds.

Key hyperparameters of RF include the number of trees (\( n_{\text{estimators}} \)), maximum tree depth (\( \text{max\_depth} \)), minimum samples required to split a node (\( \text{min\_samples\_split} \)), and the number of features considered for splitting (\( \text{max\_features} \)). Increasing \( n_{\text{estimators}} \) generally improves performance but at the expense of higher computational costs. The parameters \( \text{max\_depth} \), \( \text{min\_samples\_split} \), and \( \text{max\_features} \) regulate tree complexity, balancing bias-variance trade-offs.

For hyperparameter optimization, we employed a two-stage tuning strategy similar to that used for SVM. Table \ref{tab:comparison} also presents RF's MAE and MSE results, while Figures \ref{fig:rf_lb} and \ref{fig:rf_ub} show its confusion matrices. A more direct comparison with true PNS bounds is provided in Figures \ref{fig:rf2} and \ref{fig:rf3}. RF performs comparably to SVM on the lower bound but exhibits significantly higher accuracy on the upper bound.
\subsection{Gradient Boosting Decision Trees}
Gradient Boosting Decision Trees (GBDT) \citep{gbdt} is an ensemble learning method that builds models sequentially, with each new tree correcting the errors of its predecessors. Unlike traditional boosting, GBDT optimizes pseudo-residuals, enabling flexible loss function optimization. Simple decision trees serve as weak learners, allowing GBDT to effectively capture complex data patterns.

Key hyperparameters include the number of trees (\( n_{\text{estimators}} \)), learning rate (\( \text{learning\_rate} \)), maximum tree depth (\( \text{max\_depth} \)), and subsample ratio (\( \text{subsample} \)). The learning rate determines each tree’s contribution, while \( n_{\text{estimators}} \) and \( \text{max\_depth} \) regulate model complexity and performance.

Following the approach used for SVM and RF, we applied a two-stage tuning strategy. Again, table \ref{tab:comparison} presents the MSE and MAE results, while Figures \ref{fig:gbdt_lb} and \ref{fig:gbdt_ub} show the confusion matrices. A more direct comparison with true PNS bounds is provided in Figures \ref{fig:gbdt2} and \ref{fig:gbdt3}. GBDT demonstrates moderate performance on both the lower and upper bounds.
\subsection{Transformer}
The Transformer \citep{vaswani2017attention}, originally developed for Natural Language Processing, has expanded into Computer Vision and become a cornerstone of deep learning, particularly with the rise of large language models. Given its significant impact, this study also evaluates the Transformer for testing.

The model architecture begins with an input layer processing 15-dimensional feature vectors, followed by a linear embedding layer that projects inputs into a 64-dimensional space. Positional encoding is applied to retain feature order information, and two Transformer encoder layers with four attention heads each capture complex feature interactions. The final output is generated through a fully connected layer with a Sigmoid activation function, ensuring predictions remain within the range \([0, 1]\). Key hyperparameters include an embedding dimension of 64, four attention heads, two encoder layers, and a dropout rate of 0.1.

Similarly, table \ref{tab:comparison} presents the MSE and MAE results, while Figures \ref{fig:transformer_lb} and \ref{fig:transformer_ub} show the confusion matrices. A direct comparison with true PNS bounds is provided in Figures \ref{fig:transformer2} and \ref{fig:transformer3}. The Transformer demonstrates strong performance on the lower bound and moderate performance on the upper bound.
\subsection{Multilayer Perceptron}
MLP \citep{mlp} consists of an input layer, one or more hidden layers, and an output layer. With appropriate activation functions, it can effectively model both linear and nonlinear relationships. As a fundamental structure in deep learning, MLP holds significant representativeness, motivating its inclusion in our experiments.

A key consideration for MLP is the choice of activation function, particularly for predicting the lower bound. Since the lower bound of PNS cannot be negative, we initially selected the ReLU \citep{relu} activation function (\ref{equ:relu}). However, ReLU can lead to the loss of negatively correlated features, prompting us to adopt LeakyReLU \citep{lkrelu} (\ref{equ:lkrelu}) as a complementary solution. Furthermore, given the considerable number of zero values in the data, the non-differentiability of ReLU and LeakyReLU at \(s = 0\) imposes limitations on backpropagation. To address this, we proposed using Mish \citep{mish} (\ref{equ:mish}) as an alternative activation function. The corresponding equations are:

\begin{equation}\label{equ:relu}
    \text{ReLU}(s) = \max(0, s)
\end{equation}

\begin{equation}\label{equ:lkrelu}
    \text{LeakyReLU}(s) = 
    \begin{cases}
        s, & \text{if } s \ge 0 \\
        \alpha s, & \text{if } s < 0
    \end{cases}
\end{equation}

\begin{equation}\label{equ:mish}
    \text{Mish}(s) = s \cdot \tanh(\ln(1 + e^s))
\end{equation}

Additionally, we implemented an MLP with the architecture \( 15 \rightarrow 64 \rightarrow 32 \rightarrow 16 \rightarrow 1 \), utilizing ReLU-like functions and Sigmoid as activation functions. The model was optimized using the Adam optimizer with a learning rate of $0.01$ and trained for $1000$ epochs. 

\begin{figure*}[!htb]  
    \centering
    \begin{subfigure}[b]{0.32\linewidth}
        \centering
        \includegraphics[width=\linewidth]{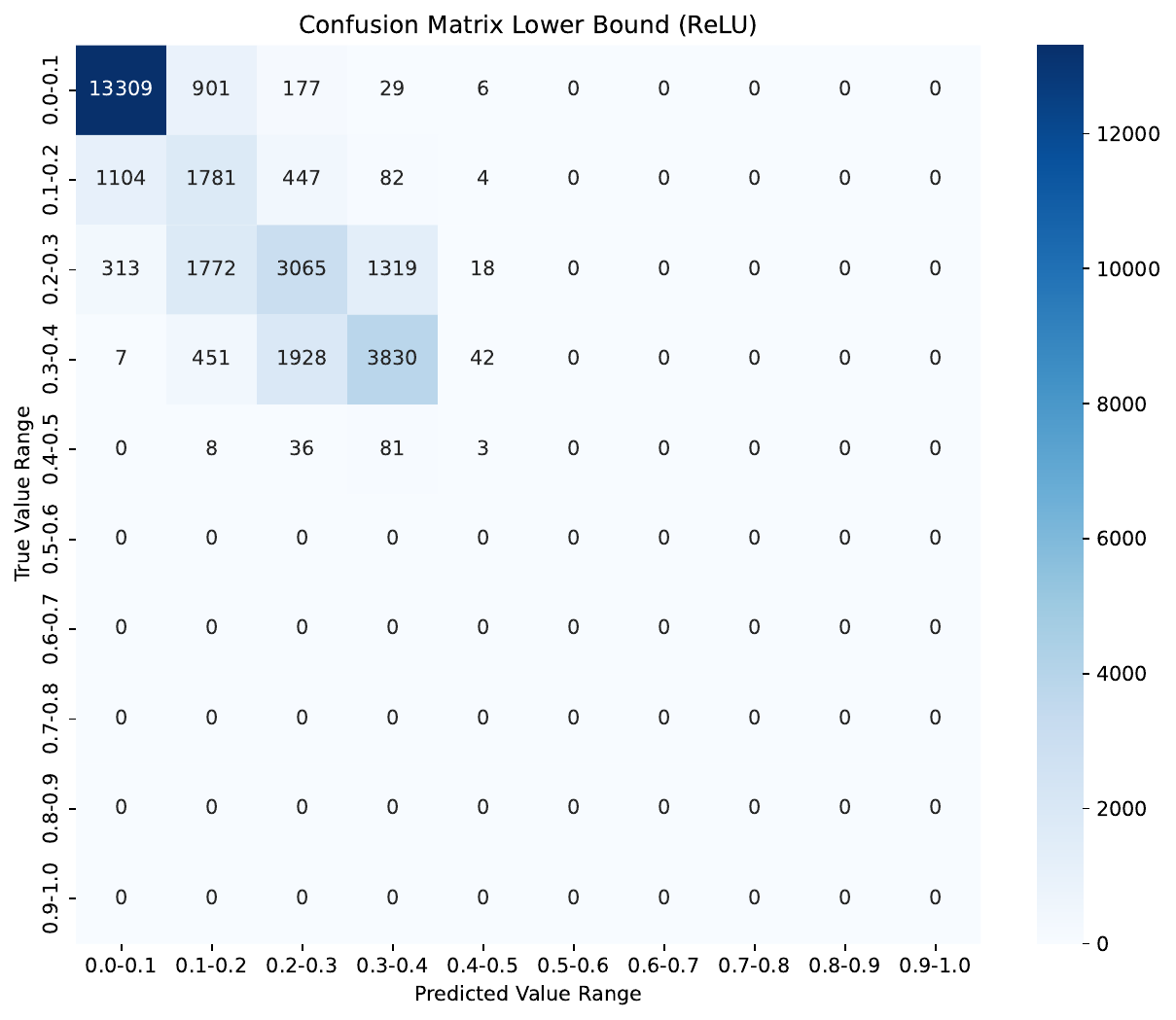}
        \caption{ReLU (Lower bound)}
    \end{subfigure}
    \hfill
    \begin{subfigure}[b]{0.32\linewidth}
        \centering
        \includegraphics[width=\linewidth]{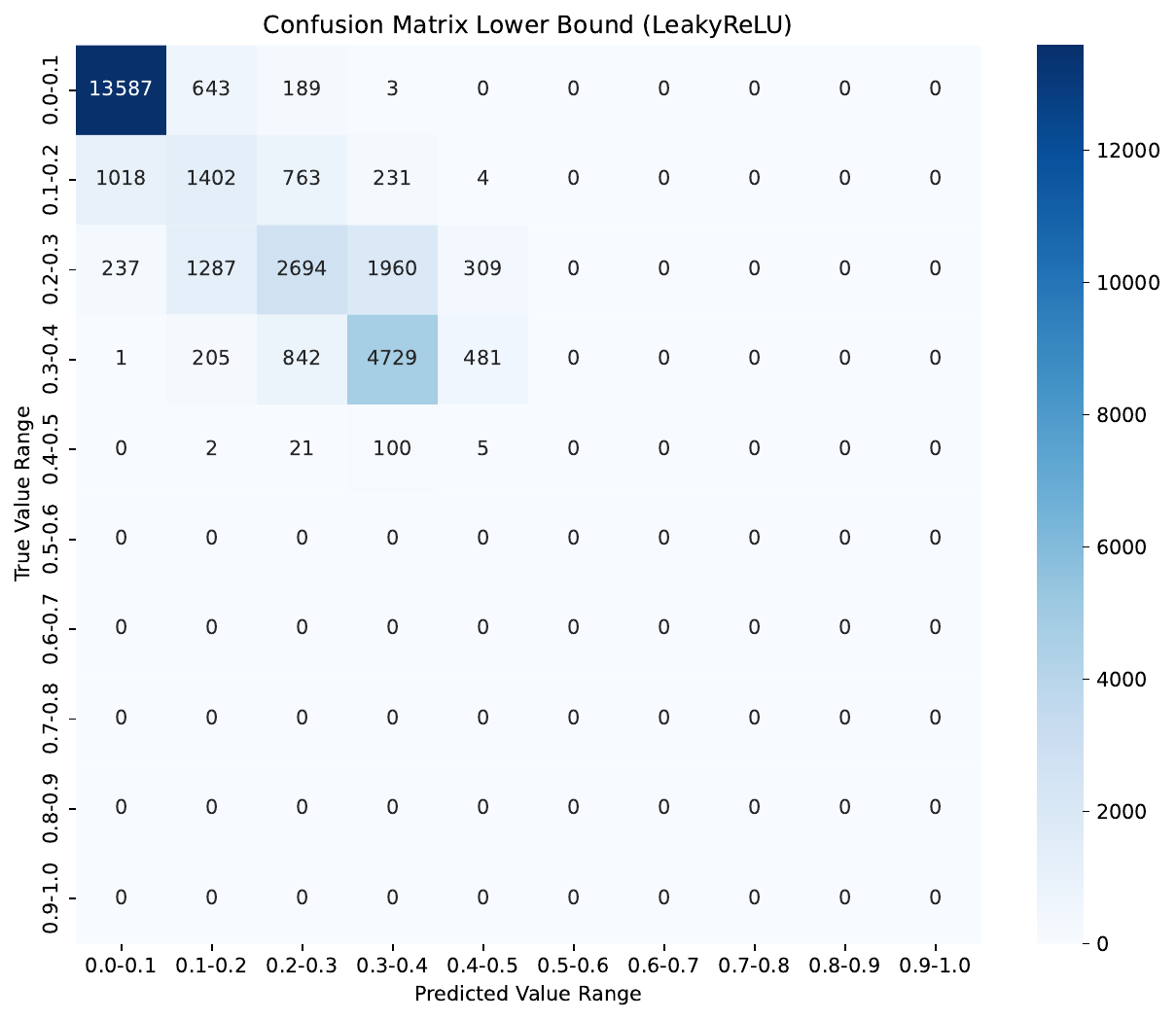}
        \caption{Leaky ReLU (Lower bound)}
    \end{subfigure}
    \hfill
    \begin{subfigure}[b]{0.32\linewidth}
        \centering
        \includegraphics[width=\linewidth]{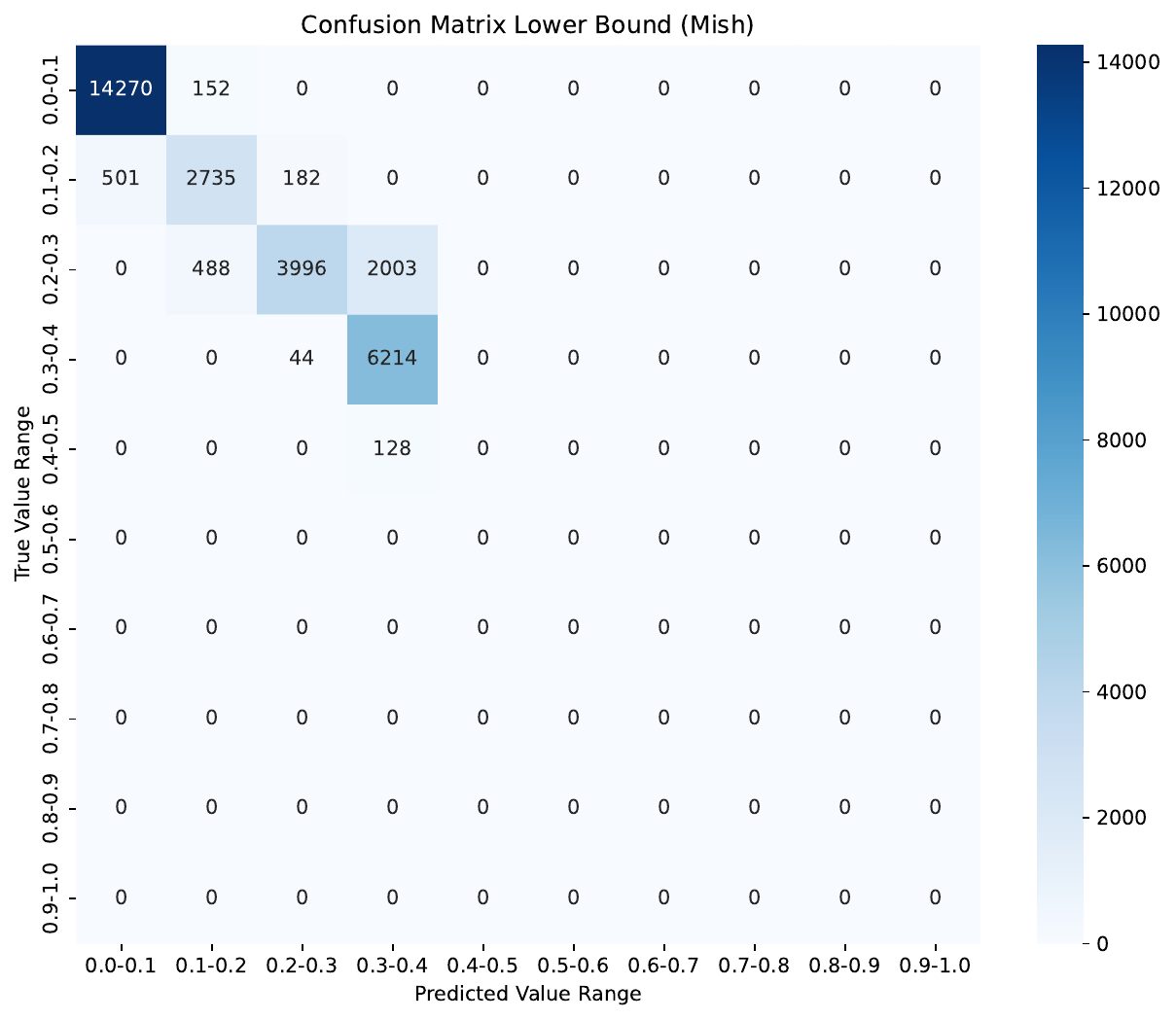}
        \caption{Mish (Lower bound)}
    \end{subfigure}

    \vspace{0.3cm} 

    \begin{subfigure}[b]{0.32\linewidth}
        \centering
        \includegraphics[width=\linewidth]{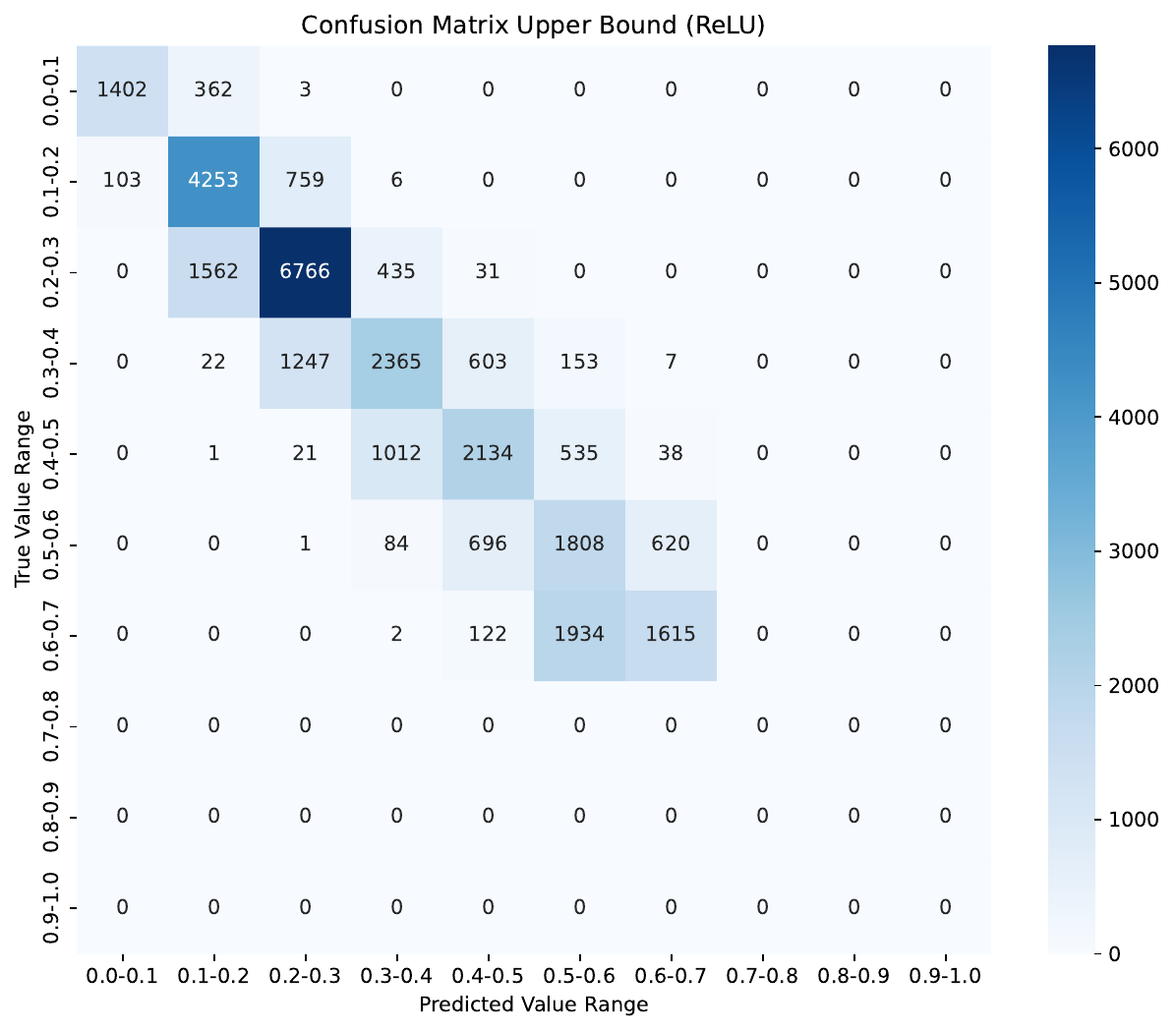}
        \caption{ReLU (Upper bound)}
    \end{subfigure}
    \hfill
    \begin{subfigure}[b]{0.32\linewidth}
        \centering
        \includegraphics[width=\linewidth]{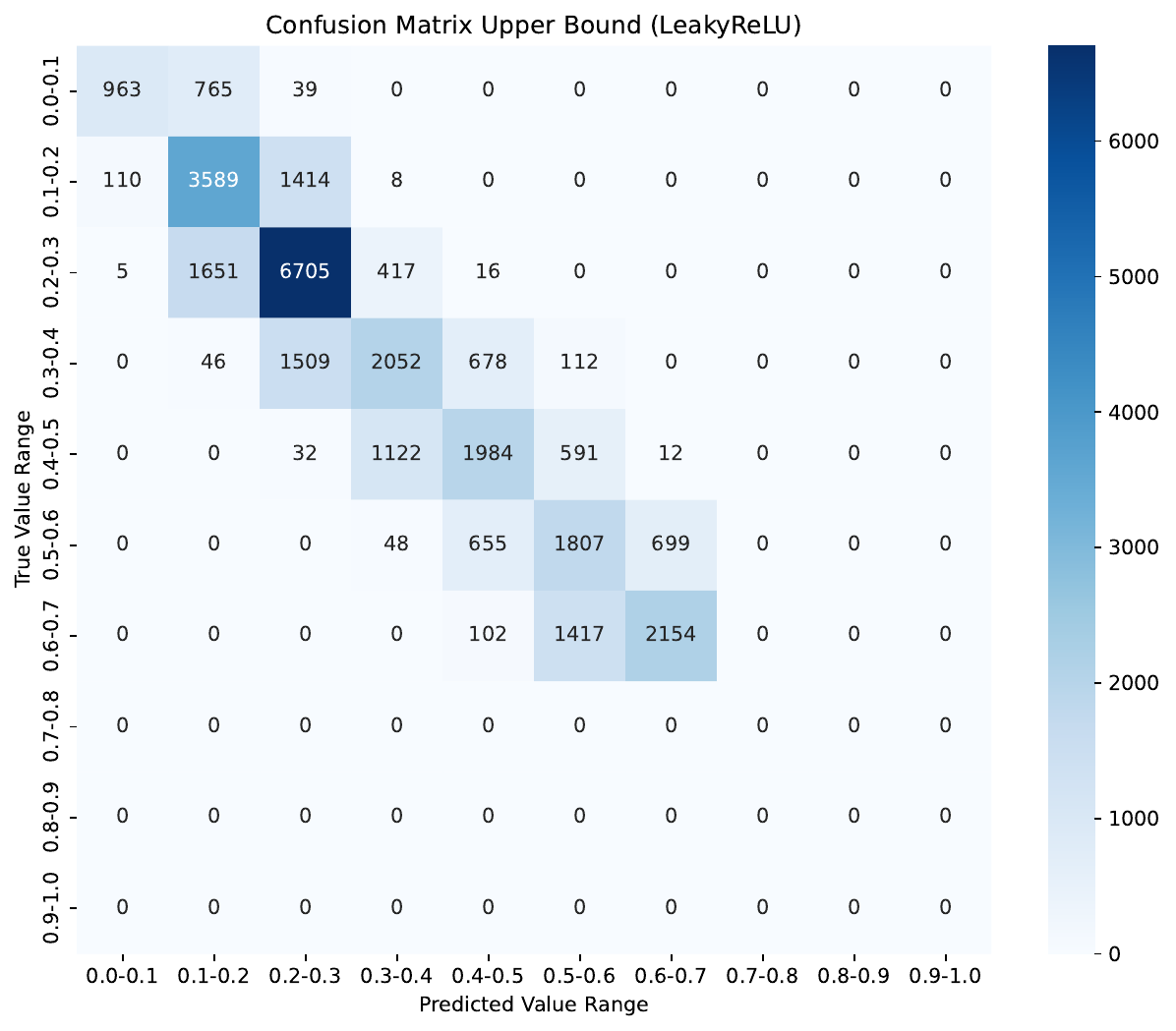}
        \caption{Leaky ReLU (Upper bound)}
    \end{subfigure}
    \hfill
    \begin{subfigure}[b]{0.32\linewidth}
        \centering
        \includegraphics[width=\linewidth]{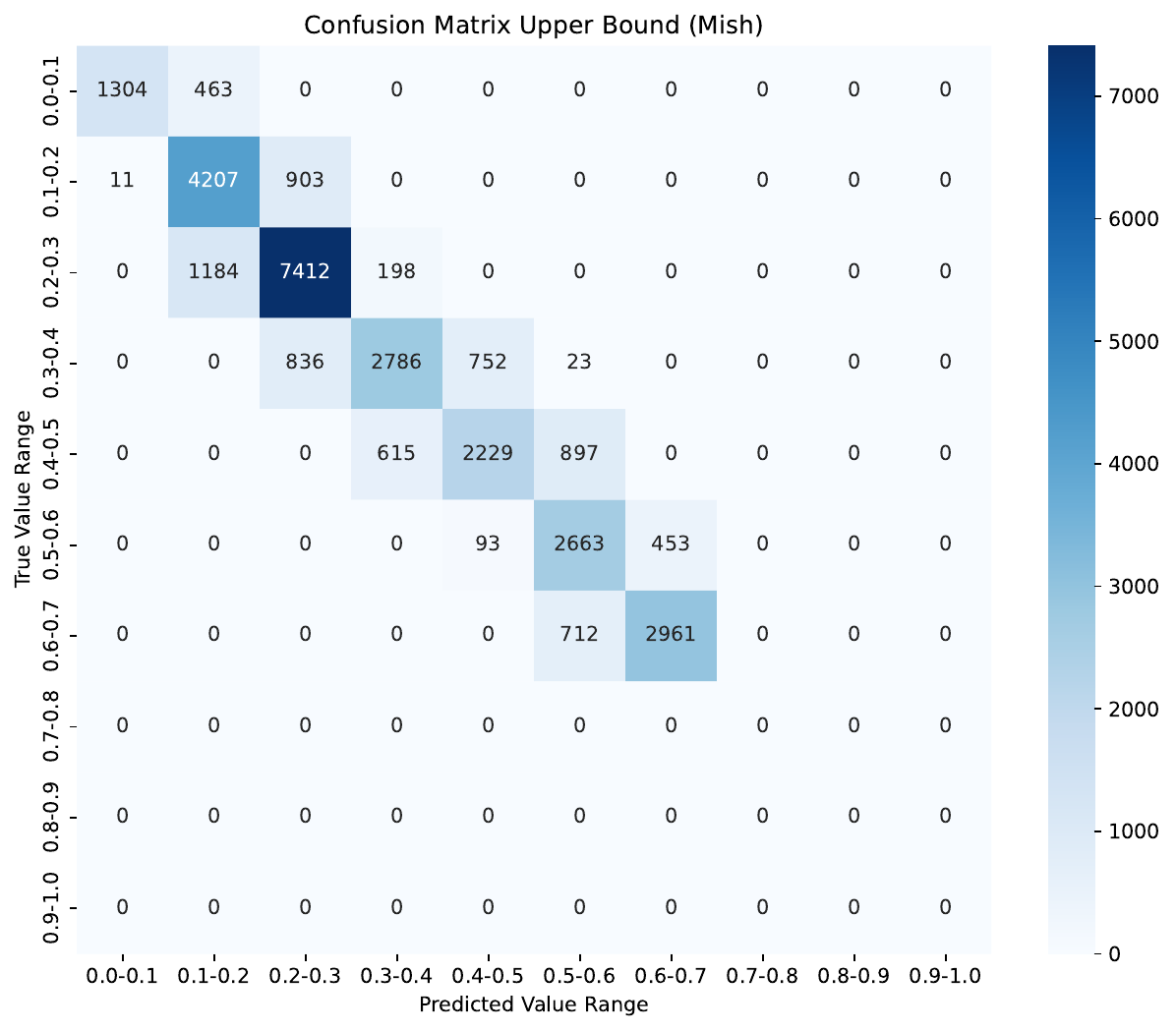}
        \caption{Mish (Upper bound)}
    \end{subfigure}

    \caption{Confusion matrices of MLP with different activation functions: ReLU, Leaky ReLU, and Mish for both lower and upper bounds.}
    \label{fig:mlp_comparison}
\end{figure*}

\begin{figure}[!htb]
    \centering
    \begin{subfigure}[b]{0.48\linewidth}
        \centering
        \includegraphics[width=\linewidth]{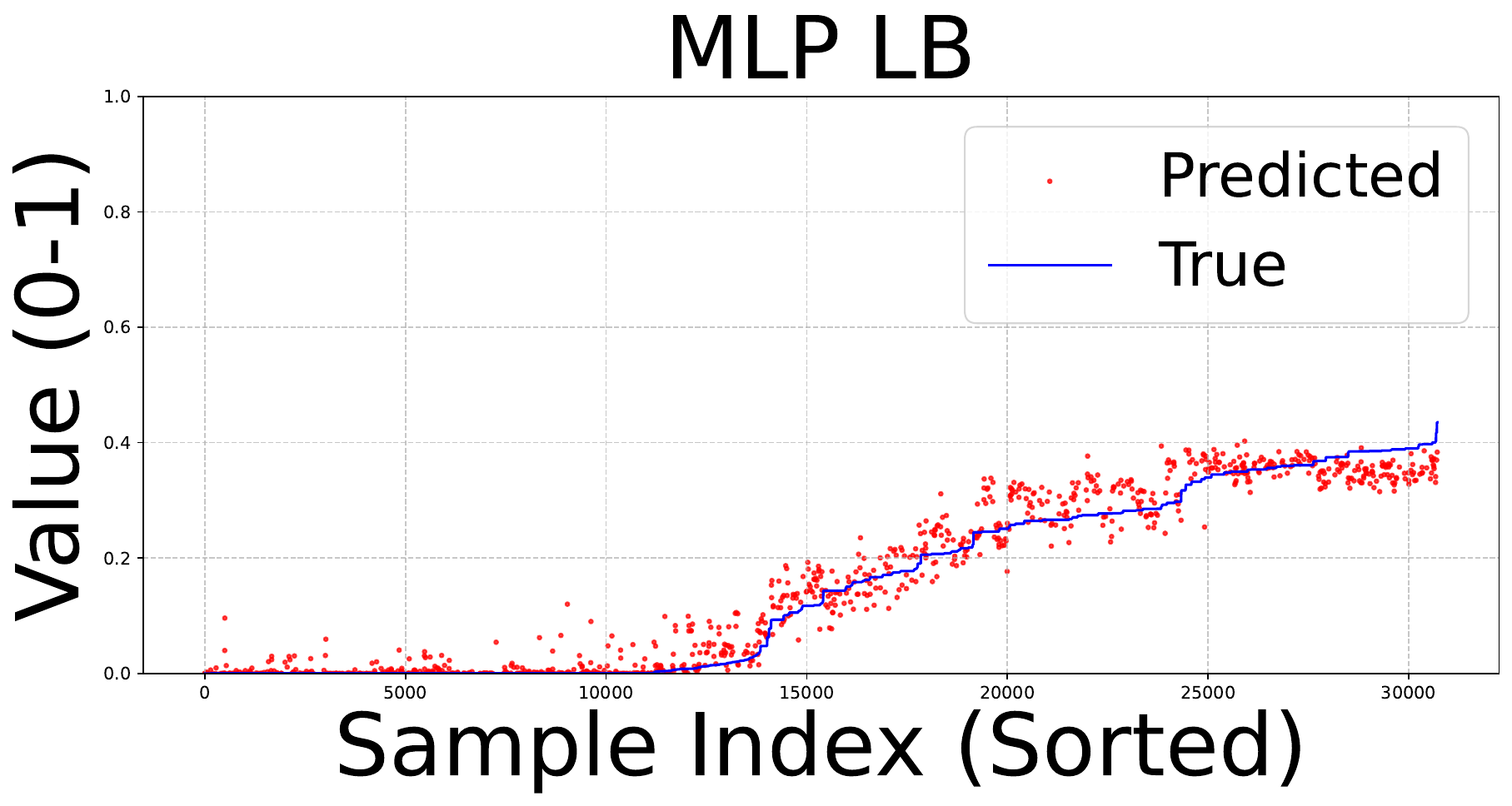}
        \caption{Lower bound.}
        \label{fig:mlp_lb}
    \end{subfigure}
    \hfill
    \begin{subfigure}[b]{0.48\linewidth}
        \centering
        \includegraphics[width=\linewidth]{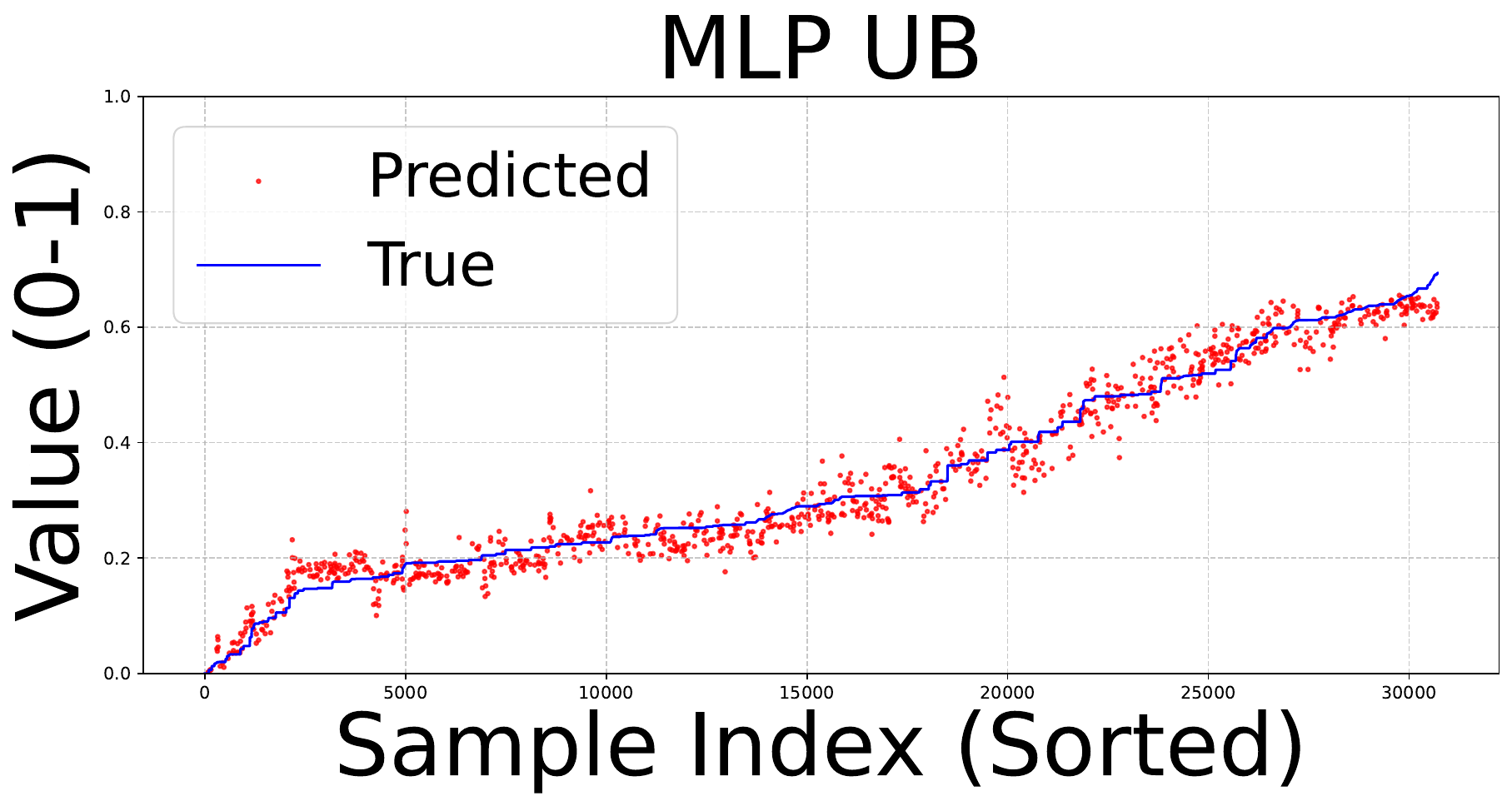}
        \caption{Upper bound.}
        \label{fig:mlp_ub}
    \end{subfigure}
    
    \caption{Comparison of MLP (Mish) for lower and upper bounds.}
    \label{fig:mlp2}
\end{figure}

Again, the final results are presented in Table \ref{tab:comparison}. With the Mish activation function, the MLP achieved an MSE of \textbf{0.0011} on the lower bound and \textbf{0.0010} on the upper bound. For MAE, it attained \textbf{0.0225} on the lower bound and \textbf{0.0247} on the upper bound. The confusion matrix is shown in Figure \ref{fig:mlp_comparison}, and a clearer comparison with the true PNS bounds is provided in Figure \ref{fig:mlp2} (Only the best performance comparisons with Mish are shown). 

Overall, MLP significantly outperformed other machine learning models, with Mish yielding the best results among the activation functions. The comparison with the true PNS bounds further confirms that MLP (Mish) provides an accurate and practical model for predicting PNS.


\subsection{Experimental Comparison}

As shown in Table \ref{tab:comparison}, MLP delivers the best overall performance. Among the other four machine learning models, SVM performs well on the lower bound but fails almost entirely on the upper bound. RF shows significantly better results, achieving acceptable performance on both bounds. Despite also being a tree-based model, GBDT underperforms compared to RF, with only a slight improvement over SVM on the upper bound. The Transformer, as an MLP-based model, outperforms the other machine learning models but still falls short of MLP’s performance.  

For MLP models, the dataset’s characteristics around zero (we will discuss these characteristics in the discussion section) lead to notable differences in activation function performance. Basic ReLU shows suboptimal performance on the lower bound, while LeakyReLU, which accounts for negative values, performs slightly better. Mish, which not only handles negative values but also ensures differentiability around zero, achieves the best results.

\begin{table}[!ht]
    \centering
-    \caption{Comparison of Model Performance} 
    \label{tab:comparison}
    \begin{tabular}{rlll}
        \toprule
        \bfseries Model & \bfseries Dataset & \bfseries MSE & \bfseries MAE \\
        \midrule
        SVM  & Lower bound & 0.0112 & 0.0868 \\
             & Upper bound & 0.0304 & 0.1527 \\
        RF   & Lower bound & 0.0116 & 0.0919 \\
             & Upper bound & 0.0205 & 0.1242 \\
        GBDT & Lower bound & 0.0159 & 0.1049 \\
             & Upper bound & 0.0261 & 0.1399 \\
        Transformer & Lower bound & 0.0030 & 0.0348 \\
             & Upper bound & 0.0156 & 0.1060 \\
        MLP(ReLU)  & Lower bound & 0.0045 & 0.0434 \\
             & Upper bound & 0.0023 & 0.0357 \\
        MLP(LeakyReLU)  & Lower bound & 0.0038 & 0.0379 \\
             & Upper bound & 0.0024 & 0.0380 \\
        MLP(Mish)  & Lower bound & \textbf{0.0011} & \textbf{0.0225} \\
             & Upper bound & \textbf{0.0010} & \textbf{0.0247} \\
             
        \bottomrule
    \end{tabular}
\end{table}
\section{Discussion}
Our study demonstrates that among common machine learning methods, the MLP (Mish) is the most effective and accurate in estimating the bounds of PNS. Below, we discuss key considerations and future directions.  

First, in complex settings—especially those involving non-binary probabilities of causation, as indicated in \citep{li2024probabilities,zhang2024causal}—the lower bounds of PNS often approach zero, further emphasizing the potential of MLP (Mish) in such scenarios.  

Second, while our model successfully addresses the challenge of predicting from approximately 2,000 subpopulations to \(2^{15}\) subpopulations, a significant issue remains: the 2,000 reliable training samples may require up to 50 million data points at the population level. This is due to data sparsity and the minimum data requirements needed to estimate PNS bounds for specific subpopulations (this paper uses 1,300 observational and experimental data points). However, collecting such a large volume of data is often impractical. Fortunately, we propose three practical approaches to data collection: instead of gathering population-level data, we can directly collect data for the 2,000 predefined subpopulations; the required 1,300 data points can be reduced to approximately 400, as suggested by \cite{li2022probabilities}, while maintaining basic accuracy; and we can rely solely on experimental data, as proposed by \cite{li2023probabilities}, or use only observational data and estimate experimental data through adjustment formulas \citep{pearl1993aspects, pearl1995causal}.  

A fundamental goal in causal inference is identifying subpopulations exhibiting desirable counterfactual behavior patterns \citep{li2019unit} to effectively guide policy-making and decision implementation. Specifically, this involves finding subpopulations with sufficiently small PNS upper bounds or sufficiently large PNS lower bounds. Therefore, accurately predicting all subpopulations is unnecessary for this task. A promising research direction is determining a minimal training set that can reliably predict these key subpopulations.  

Next, the current data-generating process consists of a relatively simple causal structure with only 20 confounders, making MLP the optimal choice. However, if the underlying causal structure were more complex or included additional confounding variables, would more sophisticated models be necessary? Since this paper focuses solely on illustrating the feasibility and potential of machine learning models for predicting causal quantities, a deeper exploration of the relationship between model complexity and causal structure remains an important direction for future research.

Finally, traditional methods for estimating probabilities of causation, such as PNS, PS, and PN, rely on direct computation using observational and experimental data within SCMs. However, they assume sufficient data for each subgroup, which is often unrealistic. Our work addresses this challenge by developing a machine learning-based framework to predict PNS for subpopulations with insufficient data, a gap not covered by traditional causal inference techniques.  

Although there are no classical methods explicitly designed for this task, we establish a robust benchmark for evaluation by leveraging the known data-generating process in our synthetic experiments. For subgroups with insufficient data, we compare our model predictions to the true PNS bounds derived from SCM equations, ensuring that our results are grounded in a well-defined standard. This explicit benchmarking demonstrates the validity of our approach and allows us to assess model accuracy relative to the traditional PNS estimation under ideal conditions.  

\section{Conclusion}
In this paper, we demonstrated that the bounds of probabilities of causation can be effectively learned and predicted using machine learning models. Specifically, we proposed five different models to predict the bounds of PNS. Experiments showed that an MLP with the Mish activation function achieved a mean absolute error of approximately 0.02 for an SCM with 15 observed and 5 unobserved confounders. Our results suggest that machine learning is a powerful tool for causal inference, particularly in real-world scenarios where direct estimation using SCM formulas is infeasible due to data limitations. Future research will explore larger datasets with more complex SCMs.

Although our study demonstrates the feasibility of machine learning for estimating probabilities of causation, we acknowledge that our experiments are based on synthetic data generated from a structured SCM. Most existing research on probabilities of causation remains theoretical, often without practical validation, despite claims of real-world applicability. Due to page limitations, we could not extend our study to real-world applications, but this remains a critical direction for future research. We believe that bridging this gap will require developing datasets from real-world causal systems where experimental and observational data can be systematically collected. Our work serves as a first step in this direction, providing a foundation for future studies to explore the practical deployment of machine learning models for causality estimation.



\clearpage
\bibliography{uai2025-template,uai2025-template/ang}
\clearpage

\onecolumn

\title{Supplementary Material}
\maketitle

\appendix
\section{The Causal Model}
The coefficients for \( M_X, M_Y \), and \( C_Y \) were uniformly generated from the range \([-1,1]\), while the parameters of the Bernoulli distribution were uniformly generated from \([0,1]\). The detailed model is as follows:
\begin{eqnarray*}
    &&\begin{cases}
        Z_i &= U_{Z_i} \text{ for } i \in \{1,...,20\},\\
        X&=f_X(M_X,U_X)\\
        &=\begin{cases}
            1& \text{ if } M_X+U_X > 0.5\\
            0& \text{ otherwise, }\\
        \end{cases}\\
        Y&=f_Y(X,M_Y,U_Y)\\
        &=\begin{cases}
            1& \text{ if } 0<C_Y \cdot X+M_Y+U_Y <1 \\
            1& \text{ if } 1<C_Y \cdot X+M_Y+U_Y <2 \\
            0& \text{ otherwise. }\\
        \end{cases}
    \end{cases}\\
&&\text{where, } U_{Z_i}, U_X, U_Y \text{ are binary exogenous variables with Bernoulli distributions.}\\
&&s.t., \\
&M_X& =
\begin{bmatrix}
Z_1~Z_2~...~Z_{20}
\end{bmatrix}\times
\begin{bmatrix}
0.259223510143\\ -0.658140989167\\ -0.75025831768\\ 0.162906462426\\ 0.652023463285\\ -0.0892939586541\\ 0.421469107769\\ -0.443129684766\\ 0.802624388789\\ -0.225740978499\\ 0.716621631717\\ 0.0650682260309\\ -0.220690334026\\ 0.156355773665\\ -0.50693672491\\ -0.707060278115\\ 0.418812816935\\ -0.0822118703986\\ 0.769299853833\\ -0.511585391002
\end{bmatrix},
M_Y =
\begin{bmatrix}
Z_1~Z_2~...~Z_{20}
\end{bmatrix}\times
\begin{bmatrix}
-0.792867111918\\ 0.759967136147\\ 0.55437722369\\ 0.503970540409\\ -0.527187144651\\ 0.378619988091\\ 0.269255196301\\ 0.671597043594\\ 0.396010142274\\ 0.325228576643\\ 0.657808327574\\ 0.801655023993\\ 0.0907679484097\\ -0.0713852594543\\ -0.0691046005285\\ -0.222582013343\\ -0.848408031595\\ -0.584285069026\\ -0.324874831799\\ 0.625621583197
\end{bmatrix}
\end{eqnarray*}
\begin{eqnarray*}
&&U_{Z_1} \sim \text{Bernoulli}(0.352913861526), U_{Z_2} \sim \text{Bernoulli}(0.460995855543),\\
&&U_{Z_3} \sim \text{Bernoulli}(0.331702473392), U_{Z_4} \sim \text{Bernoulli}(0.885505026779),\\
&&U_{Z_5} \sim \text{Bernoulli}(0.017026872706), U_{Z_6} \sim \text{Bernoulli}(0.380772701708),\\
&&U_{Z_7} \sim \text{Bernoulli}(0.028092602705), U_{Z_8} \sim \text{Bernoulli}(0.220819399962),\\
&&U_{Z_9} \sim \text{Bernoulli}(0.617742227477), U_{Z_{10}} \sim \text{Bernoulli}(0.981975046713),\\
&&U_{Z_{11}} \sim \text{Bernoulli}(0.142042291381), U_{Z_{12}} \sim \text{Bernoulli}(0.833602592350),\\
&&U_{Z_{13}} \sim \text{Bernoulli}(0.882938907115), U_{Z_{14}} \sim \text{Bernoulli}(0.542143191999),\\
&&U_{Z_{15}} \sim \text{Bernoulli}(0.085023436884), U_{Z_{16}} \sim \text{Bernoulli}(0.645357252864),\\
&&U_{Z_{17}} \sim \text{Bernoulli}(0.863787135134), U_{Z_{18}} \sim \text{Bernoulli}(0.460539711624),\\
&&U_{Z_{19}} \sim \text{Bernoulli}(0.314014079207), U_{Z_{20}} \sim \text{Bernoulli}(0.685879388218),\\
&&U_{X} \sim \text{Bernoulli}(0.601680857267), U_{Y} \sim \text{Bernoulli}(0.497668975278),\\
&&C_Y=-0.77953605542.
\end{eqnarray*}

\section{Detailed Data Generating Process}
If all 20 binary features are observable, then for a given feature set \( z = (z_1, \dots, z_{20}) \), the values of \( M_X \) and \( M_Y \) are fixed (denoted as \( M_X(z) \) and \( M_Y(z) \)). Under these conditions, the PNS, experimental distribution, and observational distribution corresponding to this set of features are:
\begin{eqnarray*}
PNS(z) &=& P(Y=0_{X=0}, Y=1_{X=1}|z)\\
&=& P(U_Y=0)\cdot T_0 + P(U_Y=1)\cdot T_1, \\
\text{where}, &T_0& =
\left \{
\begin{array}{cc}
1& \text{ if } f_Y(0,M_Y(z),0)=0 \text{ and } f_Y(1,M_Y(z),0)=1, \\
0& \text{otherwize},
\end{array}
\right.\\
&T_1& =
\left \{
\begin{array}{cc}
1& \text{ if } f_Y(0,M_Y(z),1)=0 \text{ and } f_Y(1,M_Y(z),1)=1, \\
0& \text{otherwize.}
\end{array}
\right.
\end{eqnarray*}
\begin{eqnarray*}
&&P(Y=1|do(X),z)\\
&=& P(U_Y=0)\cdot f_Y(X,M_Y(z),0) + P(U_Y=1)\cdot f_Y(X,M_Y(z),1).
\end{eqnarray*}
\begin{eqnarray*}
&&P(Y=1|X,z)\\
&=& P(U_X=0)\cdot P(U_Y=0)\cdot f_Y(f_X(M_X(z),0),M_Y(z),0)+\\
&&P(U_X=0)\cdot P(U_Y=1)\cdot f_Y(f_X(M_X(z),0),M_Y(z),1)) +\\ &&P(U_X=1)\cdot P(U_Y=0)\cdot f_Y(f_X(M_X(z),1),M_Y(z),0)) +\\ &&P(U_X=1)\cdot P(U_Y=1)\cdot f_Y(f_X(M_X(z),1),M_Y(z),1)).
\end{eqnarray*}

We assume that $15$ of the features are observable (i.e., \( Z_1, \dots, Z_{15} \)). This implies that each subpopulation, denoted as \( c = (z_1, \dots, z_{15}) \), consists of 32 possible sets of $20$ binary features. Specifically, these sets are: $s_{0}=(z_1,...,z_{15},0,0,0,0,0), s_{1}=(z_1,...,z_{15},0,0,0,0,1), s_{2}=(z_1,...,z_{15},0,0,0,1,0), ...,s_{31}=(z_1,...,z_{15},1,1,1,1,1)$.

Under this setup, we obtain the $PNS_{\text{subpopulation}}$, experimental distribution, and observational distribution for any observed subpopulation $c$ as follows:
\begin{eqnarray*}
PNS_{\text{subpopulation}}(c) &=& P(Y=0_{X=0}, Y=1_{X=1}|c)\\
&=& P(s_{0})/P(c)PNS(s_{0})+P(s_{1})/P(c)PNS(s_{1})+\\
&&P(s_{2})/P(c)PNS(s_{2})+...+P(s_{31})/P(c)PNS(s_{31})\\
&=& P(Z_{16}=0)P(Z_{17}=0)P(Z_{18}=0)P(Z_{19}=0)P(Z_{20}=0)PNS(s_{0})+\\
&&P(Z_{16}=0)P(Z_{17}=0)P(Z_{18}=0)P(Z_{19}=0)P(Z_{20}=1)PNS(s_{1})+...+\\
&&P(Z_{16}=1)P(Z_{17}=1)P(Z_{18}=1)P(Z_{19}=1)P(Z_{20}=1)PNS(s_{31}).
\end{eqnarray*}
\begin{eqnarray*}
&&P(Y=1|do(X),c)\\
&=& P(Z_{16}=0)P(Z_{17}=0)P(Z_{18}=0)P(Z_{19}=0)P(Z_{20}=0)P(Y=1|do(X),s_{0})+\\
&& P(Z_{16}=0)P(Z_{17}=0)P(Z_{18}=0)P(Z_{19}=0)P(Z_{20}=1)P(Y=1|do(X),s_{1})+\\
&& P(Z_{16}=0)P(Z_{17}=0)P(Z_{18}=0)P(Z_{19}=1)P(Z_{20}=0)P(Y=1|do(X),s_{2})+...+\\
&& P(Z_{16}=1)P(Z_{17}=1)P(Z_{18}=1)P(Z_{19}=1)P(Z_{20}=1)P(Y=1|do(X),s_{31}).
\end{eqnarray*}
\begin{eqnarray*}
&&P(Y=1|X,c)\\
&=& P(Z_{16}=0)P(Z_{17}=0)P(Z_{18}=0)P(Z_{19}=0)P(Z_{20}=0)P(Y=1|X,s_{0})+\\
&& P(Z_{16}=0)P(Z_{17}=0)P(Z_{18}=0)P(Z_{19}=0)P(Z_{20}=1)P(Y=1|X,s_{1})+\\
&& P(Z_{16}=0)P(Z_{17}=0)P(Z_{18}=0)P(Z_{19}=1)P(Z_{20}=0)P(Y=1|X,s_{2})+...+\\
&& P(Z_{16}=1)P(Z_{17}=1)P(Z_{18}=1)P(Z_{19}=1)P(Z_{20}=1)P(Y=1|X,s_{31}).
\end{eqnarray*}
The true bounds of the $PNS_{\text{subpopulation}}(c)$ can be obtained using Equations \ref{pnslb} and \ref{pnsub}, along with the above observational and experimental distributions.

\section{Code}
All code for data generation and machine learning models is available at the following anonymous link: \url{https://anonymous.4open.science/r/2025uai-ED50/}.
\end{document}